\documentclass[11pt]{article}
\iftrue
  \usepackage[margin=1in]{geometry}
  \usepackage[round,authoryear]{natbib}
  \usepackage{times}

\usepackage{amsmath,amsfonts,bm}

\def\eqref#1{equation~\ref{#1}}

\def\1{\bm{1}}

\DeclareMathAlphabet{\mathsfit}{\encodingdefault}{\sfdefault}{m}{sl}
\SetMathAlphabet{\mathsfit}{bold}{\encodingdefault}{\sfdefault}{bx}{n}

\fi

\usepackage{float}
\usepackage{placeins}
\usepackage{needspace}
\usepackage{hyperref}
\usepackage{url}
\usepackage{graphicx}

\usepackage{booktabs}
\usepackage{amsmath,amssymb}
\usepackage{microtype}
\usepackage{xcolor}
\usepackage{array}
\usepackage{flafter}
\iftrue
  \usepackage{colortbl}
  \hypersetup{hidelinks}
\fi

\newcommand{\jlens}{J-lens}
\newcommand{\jspace}{J-space}
\newcommand{\jfull}{\ensuremath{J_{\mathrm{full}}}}
\newcommand{\jread}{\ensuremath{J_{\mathrm{read}}}}
\newcommand{\resid}{\ensuremath{h}}
\newcommand{\state}{\ensuremath{s}}
\newcommand{\auc}{\textsc{AUC}}

\title{Residual Streams Read, \\
Recurrent States Remember: \\
The Global Workspace in Mamba Models}

\iftrue
\author{
Wenlong Wang\\
Fin AI Research\\
\texttt{wenlong.wang@intercom.io}
\and
Fergal Reid\\
Fin AI Research\\
\texttt{fergal.reid@intercom.io}}
\date{}
\hypersetup{
  pdftitle={Residual Streams Read, Recurrent States Remember: The Global Workspace in Mamba Models},
  pdfauthor={Wenlong Wang and Fergal Reid}}
\fi

\begin{document}

\maketitle

\begin{abstract}
Recent work proposes that verbalisable representations form a global workspace
in transformer language models: intermediate concepts occupy a shared readout
format and can be used to alter later computation. We ask what becomes of this
account in state-space language models, where information can also be carried
through recurrent state. We fit Jacobian lenses to the residual streams and
recurrent states of Mamba-1, Mamba-2, and Mamba-3, following the original
1000-prompt fitting recipe, and evaluate recovery of known intermediate concepts
on six task families. Across the tested Mamba models, combining residual and
state readouts improves intermediate-concept
recovery over the residual Jacobian lens on at least five of six task families.
On Mamba-2, the state lens alone achieves higher recovery scores than both the
residual Jacobian lens and the logit lens on all six families; a normalised
joint readout further improves on both components on five. Temporal traces and
list-memory experiments further show earlier content remaining state-readable
while its residual visibility changes. Coordinate exchange yields limited,
variable task redirection. We propose a sign-guarded steering method that
improves target top-five success over the original coordinate-exchange method
on matched verbal-report trials in all five models tested with both methods.
Recurrent state alone also supports verbal steering.
In two-hop tasks, guarded edits often output the edited concept itself rather
than the requested answer. On Mamba-3, joint residual--state edits can disrupt
successful state-only redirection. Together, these findings identify recurrent
state as a complementary carrier of workspace content, while separating
recovery, persistence, verbal steering and relational-answer success.
\end{abstract}

\section{Introduction}
\label{sec:introduction}

Consider the prompt ``The US state where the Space Needle is located shares its
northern border with \_''. Answering requires the model to identify the
intermediate concept \emph{Seattle} and then use it to produce \emph{Canada}.
The concept must therefore remain available after it is computed, so that a
later part of the computation can use it. Human cognition faces an analogous
coordination problem: information processed by one specialised system may need
to become available to others. Global workspace theories propose that selected
information enters a shared workspace from which it can be broadly accessed
and used \citep{baars1988cognitive,dehaene2001towards}. The prompt raises the
corresponding question for language models: do their intermediate concepts also
enter a common, broadly usable representational format?

Recent work investigates whether such a global workspace exists in transformer
language models \citep{gurnee2026verbalizable}. It introduces the Jacobian lens
(\jlens), which averages input--output Jacobians and uses the resulting
token-labelled directions to reveal concepts that a model is poised to
verbalise. The shared representational format exposed by this lens is called
\jspace{}. Concepts in \jspace{} are associated not only with verbal report,
but also with directed modulation, internal reasoning, flexible generalisation,
and selectivity. This provides evidence that transformer language models make
intermediate concepts broadly available, rather than merely allowing a decoder
to recover meaningful words.

The development of looped and recurrent language models creates a practical
need to extend interpretability tools beyond standard transformers. Alongside
explicit chain-of-thought generation, looped models such as Ouro perform
additional computation through repeated updates in latent space
\citep{zhu2025scaling}. These updates create intermediate representations that
are not exposed as separate steps in generated text, motivating tools that
inspect internal computation directly. Recent work extends the Jacobian lens
to looped transformers and finds that recurrence changes where workspace
content can be read and edited \citep{wang2026looped}.

State-space recurrence brings a related challenge to architectures such as
NVIDIA's Nemotron 3.5 Lightning, which combines Mamba-2 layers with attention
and mixture-of-experts layers \citep{nvidia2026nemotron35}. Whereas looped
transformers reuse layers across depth, Mamba maintains a recurrent state that
is updated at each token and carried forward through the sequence
\citep{gu2023mamba,dao2024transformers}. This state provides an additional
possible carrier for intermediate concepts, making it important to extend
interpretability tools beyond the residual stream. Does \jspace{} reside in
the residual stream, the recurrent state, or both? Does the state preserve
concepts while their residual visibility changes, and can these representations
be used to redirect subsequent computation?

We extend the \jlens{} from residual streams to recurrent SSM states and test
three increasingly strong questions:

\begin{enumerate}
    \item \textbf{Recovery:} where can a known intermediate concept be read?
    \item \textbf{Persistence:} does the state retain earlier content, and does
          observing state and residual together improve recovery?
    \item \textbf{Causal accessibility:} can lens-derived edits elicit verbal
          report and redirect relational answers?
\end{enumerate}

We study Mamba-1 2.8B, Mamba-2 2.7B, and Mamba-3 1.5B SISO and MIMO
\citep{lahoti2026mamba3}, with Pythia-2.8B \citep{biderman2023pythia} and the
Nemotron-3.5 Mamba--attention hybrid \citep{nvidia2026nemotron35} as qualified
controls. Our principal
contributions are:

\begin{itemize}
    \item To our knowledge, we are the first to identify \jspace{}
          representations in the recurrent states of selective state-space
          language models. We introduce two state \jlens{} variants: a
          full-path lens that includes propagation through future states, and a
          readout-only control restricted to the current-position output path.
          With all 63 source blocks, Mamba-2's state lens beats residual and
          logit baselines on every evaluation family.
    \item We show that state and residual readouts are complementary. A joint
          readout improves on the residual \jlens{} on at least five of six
          families for every tested Mamba model. Temporal traces and controlled
          list prompts show earlier content remaining state-readable, while also
          providing counterexamples to a universal handover story.
    \item We propose sign-guarded steering, which improves verbal-report
          success over the original coordinate-exchange method on matched
          trials across all five models tested with both methods. These gains
          in verbal report do not consistently extend to downstream two-hop
          reasoning or flexible generalisation, where outcomes depend on the
          model and intervention site.
\end{itemize}

We find evidence for verbalisable recovery and persistence in recurrent
models, with causal access depending on the task, edit construction and
intervention site. We do not infer a claim about consciousness, nor do
checkpoint comparisons isolate the effect of architecture from training, scale,
or competence.

\Needspace{10\baselineskip}
\section{Background}
\label{sec:background}

\subsection{The Jacobian lens and functional workspace claims}

Let $\resid_{\ell,t}$ be the residual activation at block $\ell$ and token
position $t$, and let $z_u$ denote the final-block residual activation before
final normalisation. For a prompt $x$, let $V_x$ be its valid fitting positions
and define $Z(x)=\sum_{u\in V_x}z_u$. The residual \jlens{} fits
\begin{equation}
    J^{h}_{\ell}
    = \mathbb{E}_{x,t}\left[
        \frac{\partial Z(x)}{\partial \resid_{\ell,t}}
      \right],
    \label{eq:res-jacobian}
\end{equation}
where prompts have equal weight and source positions $t$ are averaged within
each $V_x$. The derivative sums over current and future target positions;
causality makes terms with $u<t$ zero. Section~\ref{sec:fitting} specifies the
mask and empirical estimator. A residual activation is read by transporting it into
the final residual coordinates and applying the model's final normalisation and
unembedding. For vocabulary item $v$, the corresponding activation-space
direction is proportional to $(J^{h}_{\ell})^\top W_U[v]$, with the final norm's
gain included where required.

Unlike the logit lens, which assumes a shared residual basis at all depths, the
\jlens{} accounts for the model's average downstream transformation. Unlike a
tuned lens \citep{belrose2023tuned}, it is derived from derivatives rather than
from a supervised affine decoder. The original global-workspace study uses
the resulting token directions for both readout and causal experiments
\citep{gurnee2026verbalizable}. This distinction is central here: success as a
decoder and success as an intervention direction are related empirical
questions, not equivalent definitions.

\subsection{Residual streams and recurrent states}

For an SSM block, write its recurrent update schematically as
\begin{equation}
    \state_{\ell,t}
      = A_{\ell,t}\state_{\ell,t-1} + B_{\ell,t}x_{\ell,t},
    \qquad
    y_{\ell,t} = C_{\ell,t}\state_{\ell,t} + D_{\ell,t}x_{\ell,t}.
    \label{eq:ssm}
\end{equation}
Mamba-1 makes its selective SSM parameters input-dependent
\citep{gu2023mamba}; Mamba-2 derives a related architecture through structured
state-space duality \citep{dao2024transformers}; Mamba-3 adds a richer complex
recurrence and a MIMO formulation \citep{lahoti2026mamba3}. In each case, the
residual stream is token-local while the state compresses the prefix. This
gives two simultaneous observation points rather than two mutually exclusive
``memory systems''.

Prior work shows that some transformer interpretability methods can transfer to
recurrent language models and can benefit from their compressed state
\citep{paulo2024transfer}. Complementary work extends the \jlens{} to looped
transformers and finds workspace-like representations under weight-tied depth
recurrence, with architecture-specific read and intervention horizons
\citep{wang2026looped}. Our question is narrower and more mechanistic: does the
averaged Jacobian expose the same known intermediate concepts in residual and
state, and do its directions support the functional tests associated with the
\jspace{} account?

\section{Methods}
\label{sec:methods}

\subsection{Models and coverage}

Table~\ref{tab:models} summarises the evaluated checkpoints. The main evidence
comes from within-model comparisons; the checkpoints differ in size, training,
tokeniser, state parameterisation, and task competence. Pythia supplies a
transformer control with a residual Jacobian lens fitted using the same
1000-prompt WikiText recipe. Nemotron contains Mamba-2, MoE,
and attention blocks; its SSM state and residual stream omit the attention KV
cache, so its joint readout is incomplete.

\begin{table}[htbp]
\centering
\small
\caption{Models and paper-relevant coverage. ``State blocks'' counts source
blocks with fitted state lenses; the final block is the fit target and has no
state lens. The MIMO model has readout results but not the full state
intervention suite.}
\label{tab:models}
\begin{tabular}{lrrrrl}
\toprule
Model & Params. & Blocks & $d_{\rm model}$ & State blocks & Role \\
\midrule
Mamba-1 & 2.8B & 64 & 2560 & 63 & recurrent comparison \\
Mamba-2 & 2.7B & 64 & 2560 & 63 & principal model \\
Mamba-3 SISO & 1.5B & 24 & 2048 & 23 & recurrent comparison \\
Mamba-3 MIMO & 1.5B & 24 & 2048 & 23 & readout comparison \\
Pythia & 2.8B & 32 & 2560 & -- & transformer control \\
Nemotron-3.5 & 30B/3B active & 52 & 2688 & 23 & hybrid control \\
\bottomrule
\end{tabular}
\end{table}

\FloatBarrier
\Needspace{18\baselineskip}
\subsection{State and joint Jacobian lenses}
\label{sec:state-lenses}

For each recurrent block, flatten the post-update state
$\state_{\ell,t}$ into $d_s$ coordinates. We fit the full-path state Jacobian
\begin{equation}
    J^{s,\mathrm{full}}_{\ell}
      = \mathbb{E}_{x,t}\left[
          \frac{\partial Z(x)}{\partial \state_{\ell,t}}
        \right].
    \label{eq:state-jacobian}
\end{equation}
This includes paths through both the current output and later states of the
source block. For the readout-only control, let $y_{\ell,t}$ be its scan output
before gating and output projection, and let
$C_{\ell,t}=\partial y_{\ell,t}/\partial\state_{\ell,t}$ be the linear state
readout at the captured inputs. We retain only the path through $y_{\ell,t}$:
\begin{equation}
    J^{s,\mathrm{read}}_{\ell}
      = \mathbb{E}_{x,t}\left[
          \frac{\partial Z(x)}{\partial y_{\ell,t}}C_{\ell,t}
        \right].
    \label{eq:state-read-jacobian}
\end{equation}
The downstream derivative includes gating, output projection and all later
blocks. Thus \jread{} excludes propagation through future states of the
\emph{source block}, but retains paths to future target positions through
downstream blocks. Both variants are $d_{\rm model}\times d_s$ maps into final
residual coordinates; their transported states are decoded with the model's
final normalisation and unembedding. Fitting averages derivatives of the
frozen model and does not train a decoder on concept labels.

For Mamba-3 we express the primary lens in the current token's rotary frame;
the state and derivative use matching coordinates before averaging
(Appendix~\ref{app:state-fitting}). A fixed permutation of state coordinates before
readout supplies a control with the same dimensionality but destroyed
coordinate correspondence.

The post-update joint representation at a block is the pair
$(\resid_{\ell,t},\state_{\ell,t})$. Because both lenses map to final residual
coordinates, no additional fit is needed:
\begin{align}
    r_h &= J^h_{\ell}\resid_{\ell,t},
    &r_s &= J^s_{\ell}\state_{\ell,t},\\
    r_{\rm sum} &= r_h+r_s,
    &r_{\rm nsum} &= \frac{r_h}{\lVert r_h\rVert_2}
                 + \frac{r_s}{\lVert r_s\rVert_2}.
    \label{eq:joint}
\end{align}
We report both the plain and component-normalised sum. A pre-update convention
gives the same qualitative result except on the typo task, where the readout
token itself writes the correction into state; the main text uses post-update
readouts consistently.

\subsection{Fitting recipe}
\label{sec:fitting}

All paper-comparable residual and state lenses use 1000 WikiText-103 prompts,
truncated to 128 tokens. The reference fitting mask excludes the first 16
positions and the final position: for a tokenised prompt of length $T_x$,
$V_x=\{16,\ldots,T_x-2\}$, using zero-based indices. Both source and target
positions use this mask; excluded tokens remain in the forward computation.
For $N_{\rm p}$ fitted prompts and any of the derivative matrices
$M_{\ell}(x,t)$ above, the empirical average is
\begin{equation}
    \mathbb{E}_{x,t}[M_{\ell}(x,t)]
    \equiv \frac{1}{N_{\rm p}}\sum_{x=1}^{N_{\rm p}}
        \frac{1}{|V_x|}\sum_{t\in V_x}M_{\ell}(x,t).
    \label{eq:lens-fit-average}
\end{equation}
Target contributions are summed inside $Z(x)$, without an additional division
by the number of targets or causal source--target pairs. The target is the
final block's residual output before final normalisation, and accumulation is
performed in fp32. This matches the reference residual-lens estimator,
including for the Pythia control. Its fit uses four shards of 250 prompts, merged by their
prompt counts; the earlier non-conforming 16-prompt Pile fit is retained only
as a sensitivity comparison.

For each output coordinate, we differentiate the sum of valid target
activations with respect to captured scan outputs. Multiplication by the local
state readout gives the readout-only derivative; a reverse recurrence adds the
paths through future source-block states. Equivalent weighted sums avoid
materialising every state Jacobian. Appendix~\ref{app:state-fitting} gives the
recurrence and accumulation procedure, followed by numerical checks.

\subsection{Intermediate-recovery protocol}

We use the six official task families from the original evaluation: multihop,
multilingual mapping, order of operations, poetry, association, and typo
correction. Each supplies one or more known intermediate concepts that are
neither the prompt target nor necessarily the model's next token. Only
single-token surface forms under the model's tokeniser are scored, and coverage
is reported.

For intermediate $i$, let $r(i)$ be its best vocabulary rank over the evaluated
source blocks. We compute pass@$k$ and integrate it against $\log_{10}k$ for
$k=1,\ldots,100$, yielding a normalised \auc{} in $[0,1]$. Comparisons are
made per task family; a lens wins only when its \auc{} is higher on a majority
of the six families. This score evaluates how reliably a lens recovers known
intermediate concepts, rather than whether the model answers the task correctly.
We compare lenses using intermediate recovery; next-token prediction measures,
such as KL divergence, agreement with the model's output, and cross-entropy,
serve only as supplementary diagnostics.

Any-block readout uses every fitted source block. Causal interventions use the
fixed middle-half band: blocks 16--48 for the 64-block Mamba-1 and Mamba-2
models, and the corresponding quarter-to-three-quarter band for other depths.
We therefore report a band-restricted recovery control alongside the any-block
result.

\subsection{Persistence and causal tests}

We track known intermediate concepts across token positions and source blocks
using their vocabulary ranks under the residual and state lenses on selected
multihop prompts. On 80-word list prompts, we measure recovery of previously
presented words at later comma positions using both vocabulary-rank thresholds
and sparse non-negative concept inventories obtained by gradient pursuit. For
the reported state-inventory results, pursuit decomposes the lens-transformed
state in final-residual coordinates.

\paragraph{Intervention constructions and interfaces.}
Let $x$ be a column vector representing a block-output residual ($R$) or a
flattened recurrent state ($S$), and let $a,b$ be unit source and target lens
directions. We distinguish signed projection transfer ($N$), its sign-guarded
version ($G$), and pseudoinverse coordinate exchange ($X$):
\begin{align}
    x_N &= x + (x^\top a)(b-a), \label{eq:signed-transfer}\\
    x_G &= x + |x^\top a|(b-a), \label{eq:guarded-transfer}\\
    x_X &= x + V(P-I)V^+x. \label{eq:exchange}
\end{align}
Here $V$ has the source and target directions as columns and $P$ exchanges
their two coordinates. $X$ implements the published exchange formula
\citep{gurnee2026verbalizable}; the source's verbal-report example also
describes projection transfer. We therefore specify the operator rather than
assuming every published ``swap'' uses the same construction. Our primary
exchange convention, $X_{\rm unit}$, uses unit columns; $X_{\rm raw}$ retains
their original norms. Neither uses final-norm gamma weighting. Both use a
float64 pseudoinverse with absolute tolerance zero, relative tolerance
$10^{-10}$ and strength one. Older protocol exchanges are reported separately
in Appendix~\ref{app:interventions}.

$G$ orients the entire target-minus-source displacement. At a fixed
activation its local contrast change is non-negative:
\begin{equation}
    (x_G-x)^\top(b-a)=2|x^\top a|(1-a^\top b).
    \label{eq:local-contrast}
\end{equation}
This does not guarantee any downstream logit or task improvement. Exchange
reflects the difference coordinate when the unit columns are independent;
it is not monotonic target steering (Appendix~\ref{app:exchange-numerics}).
The operators have different native trajectories and edit magnitudes.

Edits apply at every prompt position in the fixed band. State edits use
\jfull{} and persist inside the scan; $RS$ edits both interfaces in one
forward pass. Each model uses frozen paired trials, sites and token IDs, with
three random-direction controls matched to each $G/X$ arm's realised norm at
each site. Historical $N/\jread{}$ controls and coverage are labelled
separately. In particular, Mamba-1's new comparison uses 33 state-band blocks
for verbal report but 12 sampled blocks for the suite; Mamba-2 uses all 33,
Mamba-3 all 13, and Nemotron 12 state blocks alongside 27 residual blocks.

\paragraph{Task endpoints and output audit.}
The category-report task scores the edited concept itself; the suite instead
scores the answer obtained by applying a relation or function to that concept.
These tests do not generally require extra generated reasoning tokens.
\iftrue
\begin{table}[htbp]
\centering\small
\caption{The edited concept and the scored output are task dependent.}
\label{tab:causal-tasks}
\begin{tabular}{p{.23\linewidth}p{.32\linewidth}p{.34\linewidth}}
\toprule
Task & Edited concept (example) & Scored output \\
\midrule
Verbal report & Category answer: Canada to Brazil & Brazil in next-token top five \\
Two-hop reasoning & Implicit intermediate: Brazil to Mexico & Spanish, the new intermediate's answer \\
Flexible generalisation & Explicit argument: France to Canada & Ottawa, the function's new answer \\
\bottomrule
\end{tabular}
\end{table}
\fi
The suite scores greedy output, using its saved second step when the answer
starts with whitespace. Clean original-answer competence is reported separately
from clean success on the requested swapped answer. Our post-hoc output audit
tests whether the saved first-token prediction equals the edited-concept token,
not the downstream-answer token. It also separates original answers, source
intermediates and other failures, preserving the two-step answer criterion.
This is a diagnostic of interventions, not a change to the six-family
lens-quality protocol.

\paragraph{Ablation.}
We perform multi-atom pursuit-subspace ablations in the residual stream and
recurrent state. At each probed block and token position, gradient pursuit
selects up to ten lens directions from a clean forward pass. We remove the
activation's projection onto their span in a second forward pass and report
how many initially correct answers remain correct, with random-direction
ablations as controls. Because residual activations and fitted Jacobians can
be dominated by a small number of coordinates, we also compare residual
pursuit ablations with top-coordinate and variance-matched coordinate controls.
In a modified-lens
control, we zero a Jacobian output row only when constructing intervention
directions; this does not change the model's weights or its unedited forward
pass.

\section{Recurrent states expose information missed by residual readouts}
\label{sec:recovery}

\subsection{Residual and state readouts are complementary}

Reading recurrent state improves intermediate recovery across all four tested
Mamba checkpoints (Table~\ref{tab:crossmodel}). The best evaluated state-lens
variant beats the residual \jlens{} on five of six task families for Mamba-1
and Mamba-3 SISO, all six for Mamba-2, and four for Mamba-3 MIMO. Combining
residual and state with a plain sum improves over the residual \jlens{} on
all six families for Mamba-1 and Mamba-2, and five for both Mamba-3 variants.

\begin{table}[htbp]
\centering
\small
\caption{Number of task families (out of six) on which a readout has higher
\auc{} than the within-model residual \jlens{}. ``Best state'' denotes the
highest \auc{} among the evaluated state-lens variants for each task family.
These are directional per-family comparisons, not claims that every
difference is statistically significant.}
\label{tab:crossmodel}
\begin{tabular}{lrrl}
\toprule
Model & Best state $>$ residual & Plain joint $>$ residual & State coverage \\
\midrule
Mamba-1 2.8B      & 5/6 & 6/6 & 63 source blocks \\
Mamba-2 2.7B      & 6/6 & 6/6 & 63 source blocks \\
Mamba-3 SISO 1.5B & 5/6 & 5/6 & 23 source blocks \\
Mamba-3 MIMO 1.5B & 4/6 & 5/6 & 23 source blocks \\
\bottomrule
\end{tabular}
\end{table}

Mamba-2 provides a detailed illustration
(Table~\ref{tab:mamba2-recovery}, Figure~\ref{fig:recovery}). Over all 63 source
blocks, its \jfull{} state lens beats the residual \jlens{}, logit lens and
tuned lens on all six task families (additional baselines in
Table~\ref{tab:mamba2-baselines}). The largest differences occur on
order of operations (0.795 versus 0.460 residual) and typo correction (0.706
versus 0.303); association also rises from 0.058 to 0.237. Poetry remains hard
in absolute terms: state \auc{} is 0.101, and its separation from the permuted
control is only 0.052.

\begin{table}[htbp]
\centering
\small
\caption{Mamba-2 intermediate recovery over all 63 source blocks, measured by
normalised pass@$k$ \auc{} (not task accuracy). The normalised joint readout
combines residual and state after separately normalising their transported
representations. Bold marks the best non-oracle readout in each row.}
\label{tab:mamba2-recovery}
\begin{tabular}{lrrrr}
\toprule
Task & Residual $J^h$ & State \jfull{} & Joint normalised & Permuted-state joint \\
\midrule
Multihop       & 0.475 & 0.557 & \textbf{0.679} & 0.431 \\
Multilingual   & 0.362 & 0.615 & \textbf{0.647} & 0.312 \\
Order of ops.  & 0.460 & 0.795 & \textbf{0.807} & 0.515 \\
Poetry         & 0.029 & 0.101 & \textbf{0.114} & 0.092 \\
Association    & 0.058 & \textbf{0.237} & 0.231 & 0.117 \\
Typo           & 0.303 & 0.706 & \textbf{0.714} & 0.350 \\
\bottomrule
\end{tabular}
\end{table}

The normalised joint readout scales each lens-transformed vector to unit
Euclidean norm before adding the two vectors (Equation~\ref{eq:joint}). It
exceeds both Mamba-2 components on five of six families; association is the
exception, where adding the weaker residual term
slightly dilutes the state (0.231 joint versus 0.237 state). For multihop,
combining the readouts raises \auc{} from 0.475 for residual alone and 0.557
for state alone to 0.679. Thus the state advantage over residual does not imply
that state alone is the best overall readout. The plain sum still beats the
residual on all six families but beats the state on only two. The additional
gain therefore depends on how the two readouts are combined.

\iftrue
\begin{figure}[H]
\centering
\includegraphics[width=\linewidth]{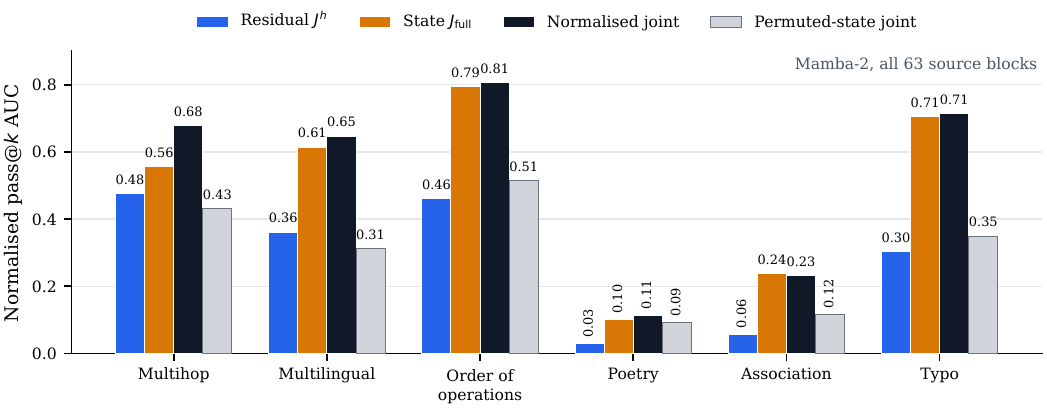}
\caption{Mamba-2 intermediate recovery over all 63 source blocks. State alone
outperforms residual alone on all six families; the normalised joint readout
exceeds both components on five, with association the exception. The
permuted-state joint control destroys coordinate
correspondence while preserving dimensionality. Values are normalised
pass@$k$ \auc{}, not task accuracy.}
\label{fig:recovery}
\end{figure}
\FloatBarrier
\fi

\subsection{Recovery across depth}

The any-block recovery scores above summarise what each interface can reveal,
but do not show where its strongest readout occurs. Across Mamba-1, Mamba-2
and Mamba-3, recovery on multihop,
multilingual and arithmetic tasks peaks earlier in state than in the residual
stream. This ordering is task dependent: typo correction often favours an
earlier residual readout.

The state advantage also persists within a shared range of intermediate depths.
Restricting both readouts to the same middle half of each network preserves
stronger state recovery on most task families in every tested Mamba checkpoint.
We next examine how state and residual readability change across token
positions as the prompt unfolds.

\section{Earlier content remains state-readable}
\label{sec:persistence}

\subsection{Following concepts across token positions}

Return to the Space Needle prompt. As Mamba-2 reads the landmark's name and
then the question about the state's northern border, we follow \emph{Seattle},
a city never named in the prompt (Figure~\ref{fig:handover}). Both the
residual and state lenses recover the city at the end of ``Needle''.

The two readouts differ in how continuously they reveal the city as the
question unfolds. At block 28, the state lens ranks Seattle first from the
end of ``Needle'' through the final prompt token. In the residual stream,
Seattle falls outside the top 100 at every source block while the model reads
``its northern border'', then reaches first rank at the final prompt position
in a later block. The state readout keeps the city visible through this gap
in the residual maps.

Other concepts follow different paths. In the emerald-birthstone question,
\emph{May} becomes prominent in the residual readout near the end of the
prompt, while remaining outside the state lens's top 100 throughout.
Mamba-2 answers with the correct month number, 5. These selected examples
show how persistence varies between concepts. We next use word lists to
measure how much earlier content remains readable across later token positions.

\iftrue
\begin{figure}[!t]
\centering
\includegraphics[width=\linewidth]{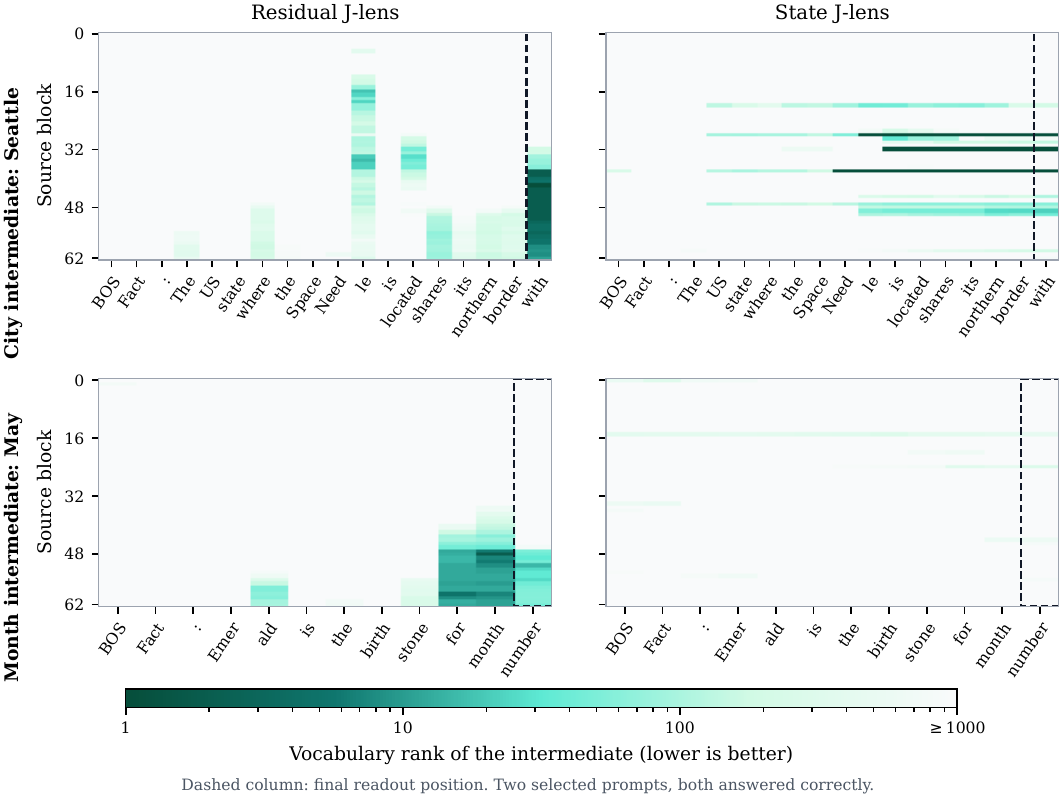}
\caption{Selected Mamba-2 rank maps across all 63 source blocks, using the
post-update convention. \emph{Seattle} remains top-ranked at state block 28
from the end of ``Needle'' through the final readout, while its residual
visibility varies. In the emerald question, \emph{May} becomes readable late
in the residual stream and remains outside the state's top 100 throughout.
Both prompts are answered correctly.}
\label{fig:handover}
\end{figure}
\fi

\subsection{Controlled evidence for earlier-token content}

Word-list prompts provide aggregate evidence that does not depend on choosing a
multihop story. After an 80-word list, we ask how many earlier words appear in
the state readout at low rank. Expanding Mamba-1 from a sparse block subset to
all 63 source blocks changes rank-based words held@5 from 0.3 to 7.5 for
\jfull{} (7.8 for \jread). Expanding Mamba-2's \jfull{} coverage from
23 to 63 source blocks raises words held@5 from 3.1 to 6.7 and non-negative
pursuit on the transported state from 10.1 to 18.6 words.
At equal coverage, Mamba-1 and Mamba-2 have similar
transported-state pursuit occupancy, 18.7 and 18.6, while their rank-based
occupancies are 7.5 and 6.7 (Table~\ref{tab:memory}).

\begin{table}[H]
\centering
\small
\caption{Word-list recovery with the \jfull{} state lens over all 63 source
blocks. Values are
last-quartile means over comma positions in 80-word lists. Pursuit uses
$k=25$ and threshold zero on the lens-transformed state in final-residual
coordinates; its occupancy and rank-based words held@5 are distinct estimators.}
\label{tab:memory}
\begin{tabular}{lrr}
\toprule
& Mamba-1 & Mamba-2 \\
\midrule
Rank words held@5 & 7.5 & 6.7 \\
Transported-state pursuit words & 18.7 & 18.6 \\
\bottomrule
\end{tabular}
\end{table}

One estimator counts top-ranked vocabulary items and the other decomposes the
transported representation. Both measure lens-recoverable content rather than
the model's total memory. The matched-coverage results remove an earlier
apparent monotonic trend across Mamba generations. State parameterisations,
tokenisers, checkpoints and competence still differ; the robust result is that
depth coverage substantially changes how much earlier content the lens reveals.

Together, Sections~\ref{sec:recovery} and~\ref{sec:persistence} establish the
positive part of the workspace extension. Verbalisable intermediates are
recoverable from recurrent state; residual and state are complementary; and
some content remains readable across later tokens. We next test whether
editing these lens-derived directions can redirect subsequent computation.

\iftrue\FloatBarrier\fi
\section{Workspace properties across residual streams and recurrent states}
\label{sec:causal}

Recovery and persistence identify recurrent state as an additional carrier of
workspace content. We now compare measured workspace properties, then examine
what the intervention endpoints reveal about access to verbal output and
two-hop answers. A readable direction, a direction that changes output, and
an edit that produces the requested relational answer are distinct findings.

\subsection{Workspace properties across models}
\label{sec:workspace-comparison}

\iftrue
Table~\ref{tab:workspace-package} brings the recurrent models together with
two transformer references. Our first-three-row measurements use one exchange
convention; the Qwen column quotes the earlier signed-clamp results of
\citet{wang2026looped}, which we did not reproduce in this study.
Category report asks whether an edited concept such as Brazil
enters the output top five; two-hop reasoning instead asks for an answer
such as Spanish after editing Brazil to Mexico; flexible generalisation asks
for Ottawa after editing France to Canada. The remaining rows measure other
workspace properties under their own protocols. These outcomes complement
the six-family intermediate-recovery \auc{}; they are not an architecture
ranking, since scale, training, tokenisation, coverage and competence differ.
Appendix~\ref{app:workspace-comparison} gives the conventions and worked examples.

\newcommand{\workspaceexchangerows}{%
\rowcolor{black!5}
\wsmeasure{Verbal report}{Target top-5} & 85/124 & 242/399 & 85/446 & 114/335 & 207/728 & 130/621 \\
\wsmeasure{Two-hop reasoning}{Requested answer} & 47/90 & 5/90 & 6/90 & 7/90 & 6/90 & 15/90 \\
\rowcolor{black!5}
\wsmeasure{Flexible generalisation}{Requested answer} & 80/192 & 13/192 & 30/192 & 12/192 & 23/192 & 29/192 \\
}

\begin{table}[htbp]
\centering
\footnotesize
\setlength{\tabcolsep}{2pt}
\setlength{\extrarowheight}{2pt}
\renewcommand{\arraystretch}{1.05}
\newcommand{\wsmeasure}[2]{\textbf{#1}\newline{\scriptsize #2}}
\newcommand{\wspair}[2]{\begingroup\setlength{\extrarowheight}{0pt}\renewcommand{\arraystretch}{1}\begin{tabular}{@{}l@{}}$R$: #1\\$S$: #2\end{tabular}\endgroup}
\caption{Workspace properties across models. Our measurements in the first
three rows use $X_{\rm unit}$ exchange: simultaneous residual--state edits ($RS$) for
recurrent models and residual edits ($R$) for Pythia. Verbal report scores
target top-five; the two suite tasks score the requested answer. Other rows
retain their separate readout or injection protocols. All Qwen3.6-27B values
are quoted from \citet{wang2026looped} and were not reproduced in this study.}
\label{tab:workspace-package}
\begin{tabular}{>{\raggedright\arraybackslash}m{0.248\linewidth}>{\centering\arraybackslash}m{0.101\linewidth}>{\centering\arraybackslash}m{0.101\linewidth}>{\centering\arraybackslash}m{0.118\linewidth}>{\centering\arraybackslash}m{0.118\linewidth}>{\centering\arraybackslash}m{0.118\linewidth}>{\centering\arraybackslash}m{0.126\linewidth}}
\toprule
& \multicolumn{2}{c}{\textbf{Transformers}} & \multicolumn{3}{c}{\textbf{SSMs}} & \textbf{Hybrid} \\
\cmidrule(lr){2-3}\cmidrule(lr){4-6}\cmidrule(lr){7-7}
\textbf{Measurement} & \shortstack{\textbf{Qwen3.6}\\\textbf{27B}$^{\dagger}$} & \shortstack{\textbf{Pythia}\\\textbf{2.8B}} & \shortstack{\textbf{Mamba-1}\\\textbf{2.8B}} & \shortstack{\textbf{Mamba-2}\\\textbf{2.7B}} & \shortstack{\textbf{Mamba-3}\\\textbf{SISO 1.5B}} & \shortstack{\textbf{Nemotron}\\\textbf{3.5}} \\
\midrule
\workspaceexchangerows
\addlinespace[4pt]
\wsmeasure{Introspection}{Injected token top-1} & 0/101 & \textbf{2/99} & \wspair{0/99}{0/99} & \wspair{0/99}{\textbf{4/99}} & \wspair{6/101}{\textbf{49/101}} & \wspair{0/99}{\textbf{93/99}} \\
\rowcolor{black!5}
\wsmeasure{Directed modulation}{Focus / control / suppress hit@5} & .38/.07/.07 & .04/.07/.04 & \wspair{.01/.05/.04}{.15/.15/.13} & \wspair{.01/.05/.04}{.15/.20/.17} & \wspair{.00/.00/.01}{.07/.09/.10} & \wspair{.02/.02/.02}{.03/.05/.02} \\
\wsmeasure{Top-down summoning}{Evoked concept} & $\approx0$ & 0 & \wspair{$\approx0$}{0} & \wspair{$\approx0$}{$\approx0$} & \wspair{0}{0} & \wspair{0}{0} \\
\rowcolor{black!5}
\wsmeasure{Language selectivity}{Explicit $-$ automatic hit@5} & +1.00 & +0.75 & \wspair{+1.00}{+0.125} & \wspair{+1.00}{+0.00} & \wspair{+0.88}{+0.375} & \wspair{+0.88}{+0.25} \\
\wsmeasure{Line-count readout}{Direct-query hit@1} & .09 & .09 & \wspair{.00}{.09} & \wspair{.00}{.00} & \wspair{.00}{.00} & \wspair{.00}{.00} \\
\addlinespace[4pt]
\rowcolor{black!5}
\wsmeasure{Rank-based words held}{Residual or $R\cup\jfull{}S$} & 0.7 & \textbf{2.1} & \textbf{7.5--8.3} & \textbf{6.7--7.2} & \textbf{5.9--6.9} & \textbf{3.1--4.7} \\
\wsmeasure{Dual-task readout}{Concept-only readable} & 6/21 & 2/21 & \wspair{0/21}{2/21} & \wspair{0/21}{5/21} & \wspair{0/21}{2/21} & \wspair{0/21}{1/21} \\
\rowcolor{black!5}
\wsmeasure{Ignition}{Countries: 10--90 width} & .10 & \textbf{.41} & \wspair{.30}{.98} & \wspair{.23}{.92} & \wspair{.94}{.95} & \wspair{\textbf{.37}}{\textbf{.85}} \\
\bottomrule
\end{tabular}
\par\smallskip
\begin{minipage}{\linewidth}
\scriptsize
$\dagger$ All values in the Qwen column are taken from \citet{wang2026looped};
none were reproduced in this study. The first three rows use signed projection
transfer. Its verbal-report cohort has 124 eligible swaps from 14 category
prompts, versus our 140 evaluation prompts per model. These are reference values,
not matched exchange comparisons (Appendix~\ref{app:workspace-comparison}).\par\smallskip
\textit{Reading the table:} $R$ = residual; $S$ = recurrent state.
Exchange uses unweighted lens vectors, a float64 pseudoinverse (atol 0,
rtol $10^{-10}$), strength 1 and persistent \jfull{} state edits.
Eligibility and coverage are model specific; Mamba-3 verbal measurements
retain a historical replay qualification (Appendix~\ref{app:exchange-numerics}).
Clean original-answer counts on two-hop are 19/90, 20/90, 25/90, 27/90 and
49/90 for Pythia, Mamba-1/2/3 and Nemotron. State introspection uses
\jread{}; other state readouts use \jread{} for Mamba-1/3 and Nemotron,
and \jfull{} for Mamba-2. Capacity uses \jfull{} and bounds $R\cup S$
from the state count to the sum of state and residual counts. Bold in these
remaining rows marks values numerically above Qwen, without a significance
claim; larger ignition widths mean less sharp transitions.
\end{minipage}
\end{table}
\FloatBarrier

In our exchange runs, the requested two-hop answer is reached on 5--15 of 90 trials and the
flexible-generalisation answer on 12--30 of 192 trials at the displayed
interfaces. Its verbal-report rate is also variable. These limited rates
characterise a particular intervention; a thresholded null does not establish
that a representation is inaccessible by another edit.

\paragraph{State-readable memory across recurrent models.}
The retained-word experiment feeds an 80-word list and, as the list grows,
counts previously presented words appearing among the lens's top-five
predictions at any evaluated block. Capacity is the mean count over the
final quarter of list positions. Appendix~\ref{app:capacity-example} shows
an actual list and the recorded counts as it is read.
The rank-based capacity lower bounds are the state counts alone: 7.5, 6.7,
5.9 and 3.1 words for Mamba-1/2/3 and Nemotron respectively. Each exceeds
both Qwen's 0.7 and Pythia's 2.1 residual-readable words. This recurring
pattern across three SSMs and a hybrid supports recurrent state as an
additional store of lens-recoverable content. These measurements describe
what the lens recovers under the stated coverage, rather than total model
memory; the ranges in the table bound the residual--state union.

\paragraph{Access to injected state content.}
Here, introspection asks whether the model can name a concept inserted into
its residual stream or recurrent state, using a prefilled reporting prompt.
We score whether the injected word is the model's top next-token prediction.
Appendix~\ref{app:introspection-example} gives the question, supplied answer
prefix and a recorded next-token prediction.
State introspection is particularly strong on Mamba-3 and Nemotron, where
the injected token is reported at top-1 on 49/101 and 93/99 trials respectively,
compared with Qwen's 0/101 and Pythia's 2/99. Mamba-1 gives 0/99 and Mamba-2
4/99, so this property is less consistent across the recurrent models than
their retained-word capacity.

The memory and injection results support distinct aspects of the workspace
account. State-readable lists show recoverable earlier content; introspection
shows externally inserted content reaching a prompted report. The latter
supplies an affirmative reporting prefix and does not test spontaneous
detection or reliable reporting of unedited internal state. It uses a direct
positive injection, separate from the exchanges and steering analysed below.
\fi

Historical signed state clamps provide a complementary positive result:
\jread{} gives 19/90 two-hop redirects on Mamba-3 and 7/90 on Mamba-1
with all 33 band blocks, against 1/90 in each matched direction control.
Flexible-generalisation counts are 19/192 and 12/192, against 2/192 and
3--4/192 controls respectively. Mamba-2's full-band signed \jfull{} clamp
gives only 1/90 despite its strong readout. Appendix~\ref{app:interventions}
retains the lens, coverage and scoring distinctions, including the separate
80/84-item protocol exchanges and multi-atom ablations.

The comparison models likewise separate recovery from intervention outcomes.
Pythia's residual \jlens{} outperforms its logit and tuned lenses on recovery,
yet its protocol coordinate exchange rarely produces the requested answer
(Appendix~\ref{app:interventions}). Nemotron shows positive intervention
outcomes despite its residual \jlens{} improving on the logit lens in only
one recovery family (Appendix~\ref{app:recovery-controls}).

\subsection{Steering residual and state representations towards verbal report}

We compare \emph{sign-guarded steering} directly with coordinate exchange,
keeping the prompts, targets and intervention sites fixed.
\iftrue
The baseline is the verbal-report row of Table~\ref{tab:workspace-package}.
\fi
Sign-guarded steering orients the edit along the target-minus-source direction
whether the activation along the source direction is positive or negative
(Equation~\ref{eq:guarded-transfer}).

\iftrue
\begin{table}[htbp]
\centering\footnotesize
\setlength{\tabcolsep}{2pt}\setlength{\extrarowheight}{2pt}
\renewcommand{\arraystretch}{1.05}
\caption{Verbal report: sign-guarded steering compared directly with Table~\ref{tab:workspace-package}. The first row repeats its verbal-report counts and model order. Both methods edit residual and state together in recurrent models and residual alone in Pythia. Success means the target enters the next-token top five.}
\label{tab:verbal-access}
\begin{tabular}{>{\raggedright\arraybackslash}m{0.248\linewidth}>{\centering\arraybackslash}m{0.101\linewidth}>{\centering\arraybackslash}m{0.101\linewidth}>{\centering\arraybackslash}m{0.118\linewidth}>{\centering\arraybackslash}m{0.118\linewidth}>{\centering\arraybackslash}m{0.118\linewidth}>{\centering\arraybackslash}m{0.126\linewidth}}
\toprule
& \multicolumn{2}{c}{\textbf{Transformers}} & \multicolumn{3}{c}{\textbf{SSMs}} & \textbf{Hybrid} \\
\cmidrule(lr){2-3}\cmidrule(lr){4-6}\cmidrule(lr){7-7}
\textbf{Intervention} & \shortstack{\textbf{Qwen3.6}\\\textbf{27B}$^{\dagger}$} & \shortstack{\textbf{Pythia}\\\textbf{2.8B}} & \shortstack{\textbf{Mamba-1}\\\textbf{2.8B}} & \shortstack{\textbf{Mamba-2}\\\textbf{2.7B}} & \shortstack{\textbf{Mamba-3}\\\textbf{SISO 1.5B}} & \shortstack{\textbf{Nemotron}\\\textbf{3.5}} \\
\midrule\rowcolor{black!5}
\textbf{Table 6 result}\newline{\scriptsize Coordinate exchange in this study} & 85/124 & 242/399 & 85/446 & 114/335 & 207/728 & 130/621 \\
\textbf{Sign-guarded steering} & -- & 298/399 & 426/446 & 333/335 & 677/728 & 621/621 \\
\bottomrule\end{tabular}\par\smallskip
\begin{minipage}{\linewidth}\scriptsize
$\dagger$ Qwen is the signed-clamp reference from \citet{wang2026looped}, not reproduced here; sign-guarded steering was not measured for Qwen. Mamba-3 retains the replay qualification in Appendix~\ref{app:exchange-numerics}. Edit magnitudes differ between the two methods.
\end{minipage}\end{table}

\FloatBarrier
\fi

Sign-guarded steering increases target-report success on all five models
tested with both methods. The largest change in success rate is on Nemotron,
from 130/621 to 621/621; Mamba-2 rises from 114/335 to 333/335.
These counts measure whether the target enters the next-token top five.
They show that the fitted directions can provide a route to verbal output
even when coordinate exchange is less successful.

The effect also appears when editing recurrent state alone, so it is not
solely a consequence of the residual intervention.
\iftrue
Appendix~\ref{app:verbal-report-details} gives the separate residual and state
results, random controls and paired uncertainty estimates. Mamba-3's
replay qualification is described in
Appendix~\ref{app:exchange-numerics}.
\fi
The methods produce different edit magnitudes, so this comparison does not
establish greater efficiency at equal dose. We next ask whether making a
concept available as output also produces the answer to a relation involving it.

\subsection{Tracing intervention outcomes in two-hop reasoning}

Reporting an edited concept is different from using it to answer a question.
After replacing Einstein with Newton, the required first-name answer is Isaac;
outputting Newton does not solve the task. We therefore inspect the saved
first-token predictions from the two-hop experiments.

\iftrue
Figure~\ref{fig:twohop-outcomes} shows two common outcomes.
\fi
Coordinate-exchange failures usually retain the unedited model's first token,
whereas sign-guarded steering often outputs the edited concept itself. This
pattern also appears on trials where the model originally answered correctly.
Stronger verbal report therefore need not translate into successful use of
the edited concept in a relation.

\iftrue
\begin{figure}[!t]
\centering
\includegraphics[width=\linewidth]{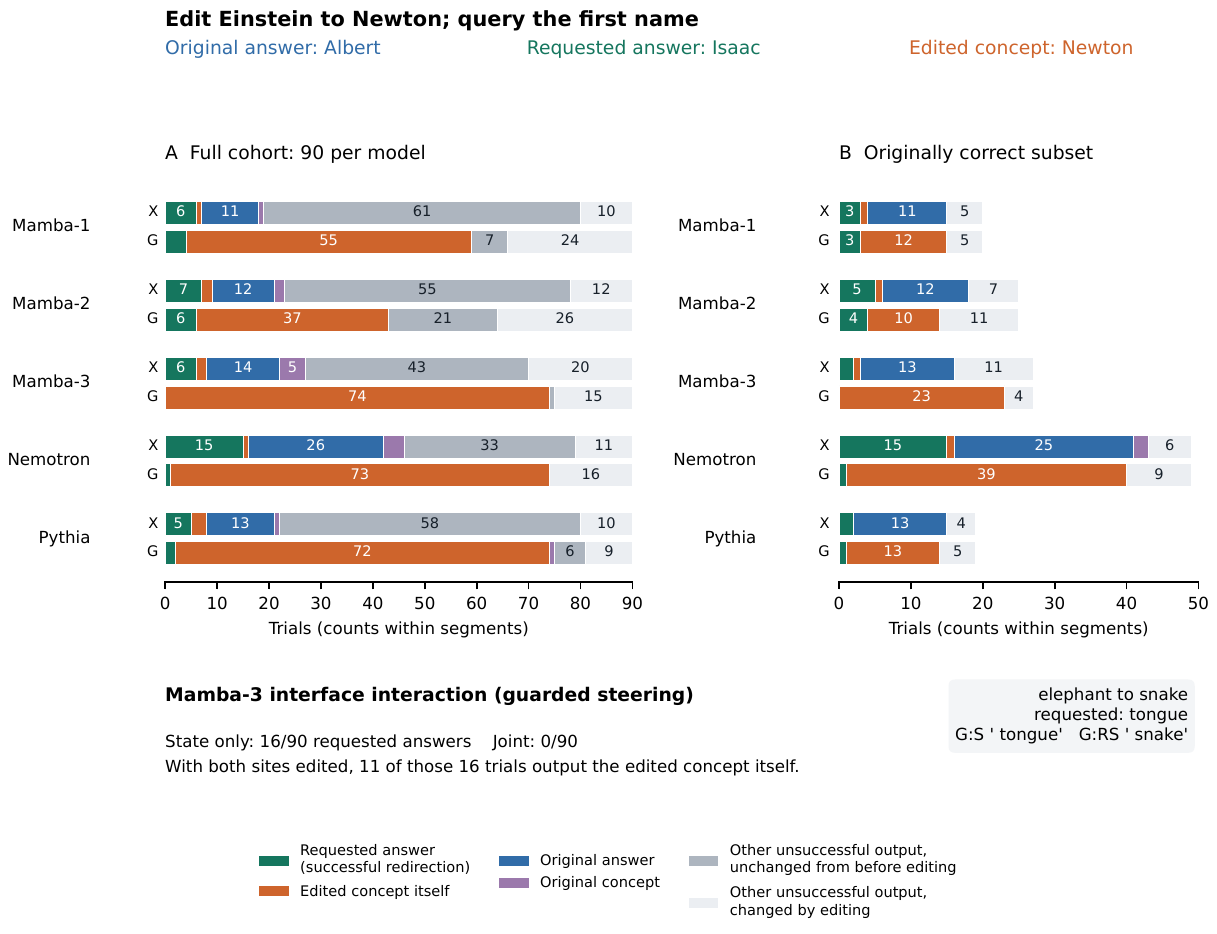}
\caption{Two-hop outputs under coordinate exchange ($X$) and sign-guarded
steering ($G$). Panel A shows all 90 trials per model; panel B shows those
answered correctly before editing. Colours distinguish the requested answer,
the edited concept and other outputs. Both methods edit residual and state
together in recurrent models, and residual alone in Pythia. The Mamba-3
callout compares state-only and joint steering. These are saved token
predictions, not complete responses; Appendix~\ref{app:twohop-output} gives
the counts, scoring rules and qualifications.}
\label{fig:twohop-outcomes}
\end{figure}
\fi

The intervention site also matters. On Mamba-3, sign-guarded steering of the
state alone produces the requested answer on 16 of 90 trials, while editing
residual and state together succeeds on none. Combining the two edits can
therefore disrupt successful redirection. These output patterns do not
identify the underlying mechanism.

\iftrue
Appendix~\ref{app:twohop-output} provides the detailed output breakdowns,
paired examples and flexible-generalisation results.
\fi

\section{Related work}
\label{sec:related}

\iftrue\begingroup\widowpenalty=10000\brokenpenalty=10000\fi
\paragraph{Verbalised representations and lenses.}
The logit lens directly unembeds intermediate residual activations, while the
tuned lens learns layer-specific affine maps \citep{belrose2023tuned}. The
Jacobian lens instead uses an average derivative to identify representations
poised to affect future verbalisation \citep{gurnee2026verbalizable}. The
original work ties these readouts to a cluster of functional workspace
properties. Our contribution is not the generic observation that decoding
does not guarantee causality; rather, we locate complementary verbalisable
content in recurrent state and test which of those functional properties
survive multiple intervention constructions.
\par
\iftrue\endgroup\fi

\paragraph{Interpretability across architectures.}
Mamba and related SSMs replace attention-based sequence mixing with selective
recurrence \citep{gu2023mamba,dao2024transformers,lahoti2026mamba3}. Prior work
has tested transformer interpretability techniques on recurrent models,
including tuned lenses and activation steering \citep{paulo2024transfer}. We
focus on the recurrent state itself, fit its full future-path Jacobian, and
compare residual, state, and joint readouts under one intermediate-recovery
protocol.

\paragraph{Global workspace as a functional analogy.}
Global workspace theories characterise access through integration, broadcast,
report, and flexible use rather than through a particular decoder. In global
neuronal workspace accounts, ignition and broadcast depend on recurrent and
long-range cortical interactions \citep{dehaene2001towards}. The original
\jspace{} study notes that a standard transformer has no such loops within a
forward pass, and hypothesises that serial processing through a deep stack of
distinct layers may play a functionally similar role
\citep{gurnee2026verbalizable}. We use workspace properties as functional
criteria for comparing residual and recurrent-state representations.

\section{Discussion}
\label{sec:discussion}

\subsection{Limitations}

Our conclusions are bounded first by experimental scope. Model comparisons do
not isolate architecture: the checkpoints differ in training data, parameter
count, tokeniser, precision, and competence. Nemotron changes attention, MoE
layers, scale, and training simultaneously, and its KV cache is unobserved. The
evaluation also scores only single-token surfaces; coverage is high but not
complete, especially for multilingual items.

A second set of limitations concerns interpretation. Some causal tests include
only items the unedited model answers correctly, so differences in baseline
capability affect the available sample and complicate cross-model comparison.
The Mamba-2 residual-Jacobian analysis in Appendix~\ref{app:geometry} measures
exact derivatives at the same token position, whereas the fitted lens also
averages effects on later positions; these analyses therefore characterise
different maps. Native edit magnitudes are not matched across operators, so
the results do not compare dose efficiency. The output partitions are post-hoc,
and saved first-token predictions do not provide complete continuations or
probability changes for the edited concept. Original-answer competence does
not establish competence on the substituted relation. Mamba-3 verbal results
also retain a one-trial historical replay discrepancy near the top-five
boundary (Appendix~\ref{app:exchange-numerics}). More broadly, the experiments
test particular functional properties associated with a workspace;
they neither establish nor refute broad claims about consciousness.

\subsection{Future work}

Context-conditioned or alternative Jacobian aggregators should be evaluated by
the same six-family recovery and intervention protocol, not by next-token
transport alone. The pure-Mamba checkpoints studied here are relatively small
(1.5--2.8B parameters) and pretrained on the Pile. As industrial-scale
pretrained or post-trained Mamba checkpoints become available, applying the
same recovery and intervention methods would enable a more informative
comparison with standard transformers. A matched-task follow-up could compare
direct naming and attribute queries for the same concept pairs, test natural
counterfactual competence, and freeze doses before comparing entity-direction,
answer-direction and relation-preserving activation edits. Saving concept and
answer probabilities plus full continuations would help distinguish the
current failure hypotheses; mediation would require its own causal test.

\section{Conclusion}

We extend the Jacobian lens to the recurrent states of selective state-space
language models, bringing an additional carrier of intermediate concepts into
the study of \jspace{}. The full-path state lens includes propagation through
future states, while a readout-only control isolates the source block's
current-position output path. Both map state into the same final-residual
coordinates as the residual lens, allowing the two representations to be
compared and combined under a common recovery protocol.

Across Mamba-1, Mamba-2 and both Mamba-3 variants, observing state alongside
the residual stream gives a fuller account of intermediate content. Joint
readouts improve on the residual \jlens{} on at least five of six task
families in every tested Mamba checkpoint. On Mamba-2, state alone exceeds
the residual and logit lenses on all six families, and the normalised joint
readout improves on both components on five. The depth profiles and temporal
experiments give this complementarity a computational context: the strongest
readouts occur at different depths, and selected concepts remain continuously
state-readable through gaps in their residual visibility. Word-list experiments
provide aggregate evidence that earlier content remains recoverable as new
tokens arrive. Together, these findings extend the workspace account to
representations carried through recurrent state across the sequence.

We also propose sign-guarded steering to direct verbalisation towards a target
concept. The method improves target top-five success over coordinate exchange
on matched trials across all five models tested with both methods. With joint
residual--state edits, success rises from 114/335 to 333/335 on Mamba-2 and
from 130/621 to 621/621 on Nemotron, demonstrating strong control over verbal
report. Recurrent state alone also supports this verbal access. These gains
are not accompanied by similarly broad improvements in downstream two-hop
reasoning or flexible generalisation. Mamba-3 makes the role of intervention location
particularly clear: state-only steering produces the requested answer on
16 of 90 trials, while simultaneous residual--state steering succeeds on none.
The benefit of combining readouts therefore coexists with interference between
edits. Understanding workspace content in recurrent models requires following
both residual and state representations through the computation, then testing
how interventions at each site, separately and together, affect its use.

\iftrue
\fi

\subsection*{Reproducibility statement}

The accompanying code specifies the fixed fitting recipe, token-surface rules,
task prompts, block bands, pass@$k$ \auc{}, and causal constructions. Raw
results are append-only and headline tables are generated from machine-readable
summaries. \iftrue
A public artefact release will provide checkpoint revisions, seeds, hardware
details, result manifests, and a stable artefact location.
\fi

\iftrue
\bibliography{main}
\bibliographystyle{plainnat}
\fi

\appendix

\section{Evaluation details}
\label{app:evaluation}

\subsection{Task families and token coverage}

The six evaluation families contain 93 multihop, 107 multilingual, 55
order-of-operations, 98 poetry, 102 association, and 96 typo prompts. Some
items contain multiple intermediates. For every intermediate we enumerate
surface forms with and without a leading space, lower-case and capitalised,
and retain only forms that encode as one token without special tokens. Mamba-2
coverage in Table~\ref{tab:mamba2-recovery} is 103/103 scored multihop
intermediates, 394/428 multilingual, 110/110 order-of-operations, 98/98 poetry,
92/92 association, and 96/96 typo.

For an item with intermediate set $I$, pass@$k$ is
\begin{equation}
  \frac{1}{|I|}\sum_{i\in I}\mathbf{1}[r(i)\leq k],
\end{equation}
averaged over items so that prompts with many intermediates do not receive
extra weight. The reported \auc{} is the trapezoidal integral of pass@$k$
against $\log_{10} k$, divided by $\log_{10}100$.

\subsection{Any-block and band-restricted recovery}

Table~\ref{tab:mamba2-band} compares recovery across all source blocks with
recovery restricted to the intervention band.
Both \jfull{} and \jread{} exceed the residual \jlens{} on all six families
within the Mamba-2 band. Within the corresponding middle-half bands for
Mamba-1 and both Mamba-3 checkpoints, both state lenses exceed the residual
on five families, with typo correction the exception. The joint readout is
not uniformly best within the Mamba-2 band.

\begin{table}[htbp]
\centering
\small
\caption{Mamba-2 state and residual \auc{} over all source blocks and over the
fixed intervention band.}
\label{tab:mamba2-band}
\begin{tabular}{lrrrr}
\toprule
& \multicolumn{2}{c}{Any block (0--62)} & \multicolumn{2}{c}{Band (16--48)}\\
Task & State & Residual & State & Residual \\
\midrule
Multihop      & 0.557 & 0.475 & 0.444 & 0.313 \\
Multilingual  & 0.615 & 0.362 & 0.601 & 0.276 \\
Order of ops. & 0.795 & 0.460 & 0.589 & 0.266 \\
Poetry        & 0.101 & 0.029 & 0.078 & 0.014 \\
Association   & 0.237 & 0.058 & 0.196 & 0.051 \\
Typo          & 0.706 & 0.303 & 0.616 & 0.258 \\
\bottomrule
\end{tabular}
\end{table}

\subsection{Additional recovery baselines and comparison models}
\label{app:recovery-controls}

Table~\ref{tab:mamba2-baselines} gives the logit and tuned baselines and the
permuted-state control for Mamba-2. The latter permutes the state coordinates
before the state lens; it is distinct from adding a permuted-state component
to the residual readout in Table~\ref{tab:mamba2-recovery}.

\begin{table}[htbp]
\centering
\small
\caption{Mamba-2 normalised pass@$k$ \auc{} over all 63 source blocks.
The \jfull{} state lens exceeds both residual baselines on all six families.}
\label{tab:mamba2-baselines}
\begin{tabular}{lrrrr}
\toprule
Task & State \jfull{} & Logit & Tuned & Permuted state \\
\midrule
Multihop & 0.557 & 0.379 & 0.441 & 0.021 \\
Multilingual & 0.615 & 0.449 & 0.446 & 0.056 \\
Order of ops. & 0.795 & 0.647 & 0.666 & 0.164 \\
Poetry & 0.101 & 0.049 & 0.036 & 0.049 \\
Association & 0.237 & 0.027 & 0.036 & 0.022 \\
Typo & 0.706 & 0.292 & 0.279 & 0.040 \\
\bottomrule
\end{tabular}
\end{table}

Pythia's conforming \jlens{} exceeds the logit and tuned lenses on all six
recovery families, both over all source layers and within the intervention
band (Table~\ref{tab:pythia-recovery}). Its any-layer \auc{} values differ
from the earlier 16-prompt Pile fit by at most 0.016, retaining the six-family
ordering. The new checkpoint revision
is \texttt{2a259cdd}; the old fit did not record its revision. Fitting corpus
and prompt count also changed together, so this comparison does not isolate
their individual effects. The tuned baseline uses the existing Pile-validation
fit. The new Jacobian lens covers source layers 0--30 of the 32-block model,
with band 8--24. Its band-restricted \auc{} values are 0.404, 0.315, 0.663,
0.041, 0.008 and 0.350 in table order, exceeding logit and tuned on every family.

\begin{table}[htbp]
\centering
\small
\caption{Pythia-2.8B recovery \auc{} over all 31 source layers. The primary
J-lens uses the conforming 1000-prompt WikiText recipe; the old 16-prompt
Pile lens is a non-conforming sensitivity comparison.}
\label{tab:pythia-recovery}
\begin{tabular}{lrrrr}
\toprule
Task & WikiText J-lens & Old Pile J-lens & Logit & Tuned \\
\midrule
Multihop & 0.470 & 0.474 & 0.296 & 0.353 \\
Multilingual & 0.324 & 0.333 & 0.214 & 0.217 \\
Order of ops. & 0.736 & 0.748 & 0.583 & 0.674 \\
Poetry & 0.063 & 0.078 & 0.018 & 0.031 \\
Association & 0.008 & 0.014 & 0.001 & 0.002 \\
Typo & 0.377 & 0.369 & 0.287 & 0.239 \\
\bottomrule
\end{tabular}
\end{table}

\Needspace{4\baselineskip}
Nemotron's residual \jlens{} exceeds its logit lens on only one recovery
family. Its plain residual--state sum improves on the residual in four of six
families under the post-update convention, but omits the attention KV cache.
These checkpoint comparisons do not isolate the contribution of attention.

\subsection{Recovery cost of removing lens output row 2107}

The modified Mamba-2 lens changes the readout map and the directions derived
from it, without changing model weights or unedited activations. Three
family-level losses exceed 0.02 \auc{} (Table~\ref{tab:row2107}); exchange
counts are unchanged. This is output-row removal, distinct from removing
an input column of the Jacobian.

\begin{table}[htbp]
\centering
\small
\caption{Mamba-2 residual J-lens recovery, original and with output row 2107
zeroed. Changes use unrounded \auc{} values.}
\label{tab:row2107}
\begin{tabular}{lrrr}
\toprule
Task & Original & Row zeroed & Change \\
\midrule
Multihop & 0.475 & 0.443 & $-0.0322$ \\
Multilingual & 0.362 & 0.335 & $-0.0267$ \\
Order of ops. & 0.460 & 0.417 & $-0.0436$ \\
Poetry & 0.029 & 0.045 & $+0.0157$ \\
Association & 0.058 & 0.056 & $-0.0013$ \\
Typo & 0.303 & 0.304 & $+0.0008$ \\
\bottomrule
\end{tabular}
\end{table}
\FloatBarrier

\section{Intervention conventions and denominators}
\label{app:interventions}

Three comparisons must remain distinct: the legacy residual protocol exchange,
historical signed workspace clamps ($N$), and the new paired workspace
$N/G/X$ comparison. The legacy residual exchange uses gamma-weighted directions
and its original numerical implementation; the protocol excludes items lacking
single-token surfaces and reports 80--84 scorable items depending on model and
tokeniser. The paired comparison uses 90 two-hop items, unweighted directions
and reference float64 exchange at both interfaces. Prompt rendering, scoring,
lens choice and coverage are retained with each result. The 22/84 state-clamp
result below uses $N/\jread{}$; it is not a new $X/\jfull{}$ suite result.

For Mamba-2, the historical signed workspace result is 1/1/1 residual/state/both clamp
hits out of 90, with 25 original answers correct. Its residual protocol
exchange is 10/80 overall and 7/23 conditioned on clean correctness
(Table~\ref{tab:mamba2-causal-rates}). Mamba-3's workspace \jread{} result is
19/90 with 27 clean original answers correct. Its separate protocol comparison
uses 84 scorable items and 29 clean-correct items
(Table~\ref{tab:mamba3-causal}). All intervals below are Wilson 95\%
intervals in percentage units; denominators count evaluation items.
Random-control seeds reuse items and are reported separately, without pooling
denominators. Intervals describe individual rates, not paired treatment effects
or tests of differences between models; they do not account for dependence
between related prompts within a task category.

For Pythia's conforming fit, protocol coordinate exchange succeeds on 3 of
80 scorable items (95\% interval 1.3--10.5\%), including none of the 19 items
the clean model initially answers correctly (0--16.8\%). The earlier
16-prompt Pile fit gave 2 of 80.

\begin{table}[htbp]
\centering
\small
\caption{Persistent state clamps: requested swap counts out of 90, with Wilson
95\% intervals in brackets (percentage units). ``Clean'' counts correct original
answers. The sparse Mamba-2 lens has 23 source blocks, 12 in the intervention
band. Controls match Mamba-1/3 \jread{} and full-coverage Mamba-2 \jfull{};
each of three random seeds and the shuffled-pair run gives the stated count.}
\label{tab:state-clamp-controls}
\begin{tabular}{lrrrl}
\toprule
Model & Band blocks & \jfull{} & \jread{} & Clean \\
\midrule
Mamba-1 & 33 & 1 [0.2, 6.0] & 7 [3.8, 15.2] & 20 \\
Mamba-2 & 33 & 1 [0.2, 6.0] & -- & 25 \\
Mamba-2, sparse & 12 & 1 [0.2, 6.0] & 1 [0.2, 6.0] & 25 \\
Mamba-3 SISO & 13 & 12 [7.8, 21.9] & 19 [14.0, 30.6] & 27 \\
\midrule
Random / shuffled & matched & \multicolumn{2}{c}{1 [0.2, 6.0] per run} & -- \\
\bottomrule
\end{tabular}
\end{table}

\begin{table}[htbp]
\centering
\small
\caption{Mamba-2 coordinate exchange and ten-atom pursuit ablations.
Exchange success and retained correct answers are different outcomes.
The two coordinate controls match the ablation comparison in Figure~\ref{fig:causal}.}
\label{tab:mamba2-causal-rates}
\begin{tabular}{lrr}
\toprule
Construction / outcome & Count & 95\% interval (\%) \\
\midrule
Residual exchange, all scorable & 10/80 & [6.9, 21.5] \\
Residual exchange, given clean correct & 7/23 & [15.6, 50.9] \\
Original pursuit, retained & 22/37 & [43.5, 73.7] \\
Row-zeroed pursuit, retained & 12/37 & [19.6, 48.5] \\
Coordinates 2--11, retained & 22/37 & [43.5, 73.7] \\
Random coordinates 2--21, retained & 26/37 & [54.2, 82.5] \\
State pursuit, all 33 band blocks, retained & 37/37 & [90.6, 100.0] \\
\bottomrule
\end{tabular}
\end{table}

\begin{table}[htbp]
\centering
\small
\caption{Mamba-3 protocol interventions using raw lens vectors, at blocks
6--18 and every prompt position. Counts include Wilson 95\% intervals in
brackets (percentage units). The final column applies both residual and
state edits; for clamp rows, both use clamp transfer.}
\label{tab:mamba3-causal}
\setlength{\tabcolsep}{4pt}
\begin{tabular}{lrrr}
\toprule
Construction & State, all & State, clean correct & Both, all \\
\midrule
\jread{} exchange & 1/84 [0.2, 6.4] & 0/29 [0.0, 11.7] & 6/84 [3.3, 14.7] \\
\jread{} clamp & 22/84 [18.0, 36.5] & 14/29 [31.4, 65.6] & 15/84 [11.1, 27.4] \\
\jfull{} clamp & 13/84 [9.3, 24.7] & 8/29 [14.7, 45.7] & 14/84 [10.2, 26.1] \\
\bottomrule
\end{tabular}
\end{table}

For flexible generalisation, the real unit-strength state clamps give
19/192 on Mamba-3 (6.4--14.9\%), 12/192 on Mamba-1 (3.6--10.6\%) and
3/192 on Mamba-2 (0.5--4.5\%). Random and shuffled controls give 2/192
(0.3--3.7\%) on Mamba-3/2 and 3--4/192 on Mamba-1 (0.5--4.5\% for
3/192; 0.8--5.2\% for 4/192). Mamba-1 at twice the clamp strength gives
0/192 (0--2.0\%).
\iftrue\FloatBarrier\fi

\begin{table}[H]
\centering
\small
\caption{Historical signed clamp ($N$) by intervention location: requested swaps in
the 90-item workspace suite. State edits use \jfull{} and persist through the
recurrence; all edits apply at every prompt position. Mamba-1 edits residual
blocks 16--48 and 12 sampled state blocks within that band; Mamba-2 edits all
33 blocks in the band for both representations. Mamba-3 is the SISO checkpoint,
with both representations edited at blocks 6--18. Comparisons within each row
use the same prompts and scoring.}
\label{tab:clamps}
\begin{tabular}{lrrr}
\toprule
Model & Residual only & State only & Both \\
\midrule
Mamba-1 & 10/90 & 1/90 & 11/90 \\
Mamba-2 & 1/90 & 1/90 & 1/90 \\
Mamba-3 & 3/90 & 12/90 & 14/90 \\
\bottomrule
\end{tabular}
\end{table}

Mamba-1's simultaneous run uses only 12 sampled state blocks alongside all
33 residual-band blocks. Its 11/90 joint count, versus 10/90 for residual
alone, does not test combining the residual edit with the stronger full-band
\jread{} state clamp. These coverage and lens distinctions limit comparisons
between intervention locations.

\subsection{Mamba-2 residual interventions and geometry diagnostics}
\label{app:mamba2-diagnostics}

Under the residual protocol's pseudoinverse coordinate exchange, Mamba-2
produces the requested answer on 10/80 scorable items, including 7/23 items
the clean model answers correctly. The workspace clamp and protocol exchange
alter different geometric quantities: nearly parallel token directions can
produce a small clamp displacement while their pseudoinverse coordinates
remain exchangeable. They also differ in prompt rendering and scoring.
This geometric observation suggests why the constructions can behave
differently, but does not establish the cause of Mamba-2's weak signed-clamp task response. Guarded verbal steering
shows that this weakness is not a general absence of causal access to state.

At full intervention-band coverage, Mamba-2's historical signed-clamp run gives
residual/state/both counts of 1/1/1 out of 90 for probe swaps and 2/3/2 out
of 192 for flexible generalisation. Its ten-atom state pursuit ablation
retains all 37 initially correct answers. The positive residual interventions
below therefore do not demonstrate causal use of the readable state directions.

Massive residual coordinates differ sharply across the tested models. At the
middle of the intervention band, Mamba-2 coordinate 2107 and Mamba-3 SISO
coordinate 1345 have respectively 128 and 172 times the median coordinate
RMS. Across the band, projecting out these coordinates removes on average
85\% and 94\% of residual squared norm and leaves none of the initially
correct answers unchanged (0/37 and 0/32). By contrast, the largest
coordinates in Mamba-1 and Nemotron account for approximately 20\% of residual
squared norm; removing them retains 22/27 and 43/47 answers. A single dominant
residual channel is therefore shared by Mamba-2 and Mamba-3, but is not
universal across the tested recurrent and hybrid models.

\begin{figure}[H]
\centering
\includegraphics[width=\linewidth]{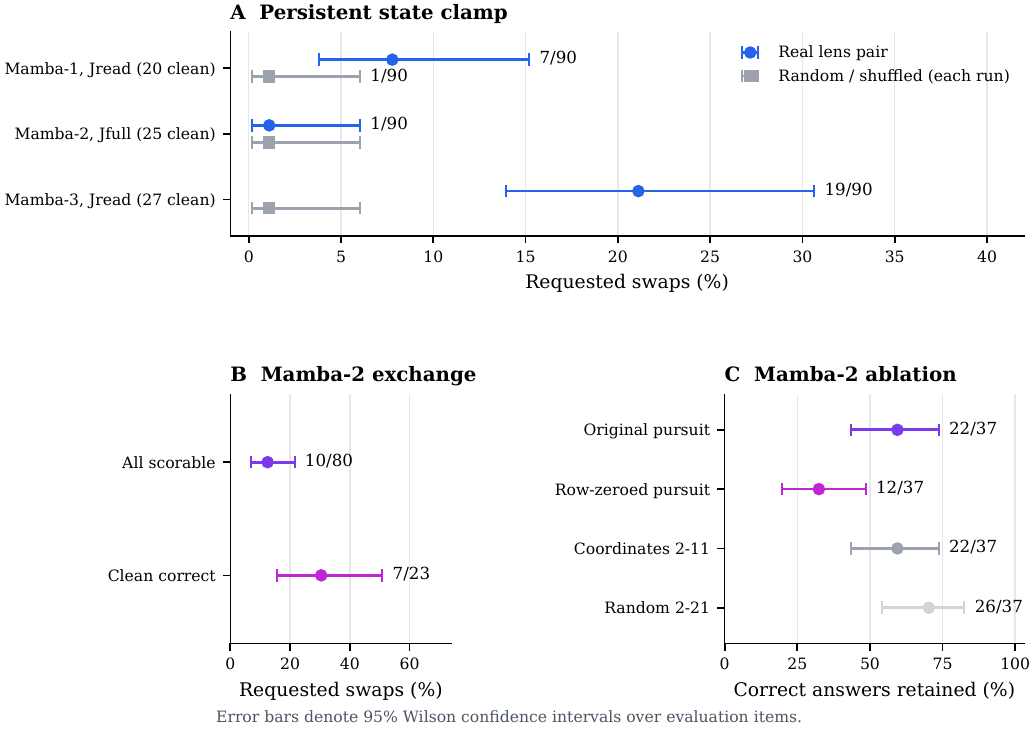}
\caption{Construction-specific causal evidence. Panel A compares persistent
state clamps and matched direction controls in the common 90-item workspace
suite; the three random seeds and shuffled pairs each yield 1/90 and share a
single plotted control marker. Panel B reports Mamba-2's
residual coordinate exchange under the separate protocol. Panel C compares
Mamba-2 pursuit ablations and variance-matched coordinate controls; fewer
retained answers indicate a larger effect. Denominators and constructions
are shown explicitly and must not
be pooled into one causal score.}
\label{fig:causal}
\end{figure}

A related diagnostic uses multi-direction ablation. In Mamba-2, the
massive coordinate also appears as a dominant output row of the fitted lens:
row 2107 contributes 12--34\% of its Frobenius norm and gives many token
directions a shared, nearly parallel component. As a diagnostic, we zero this
lens row---without changing the model or its activations---before constructing
a ten-direction pursuit subspace. Ablating the resulting subspace leaves only
12 of 37 initially correct answers unchanged, compared with 22 of 37 for the
original pursuit subspace and 22 or 26 of 37 for variance-matched coordinate
controls (Figure~\ref{fig:causal} and
Table~\ref{tab:mamba2-causal-rates}). The completed six-family protocol now
shows a recovery cost: \auc{} falls by 0.0322 on multihop, 0.0267 on
multilingual and 0.0436 on order of operations, while rising by 0.0157 on
poetry (Table~\ref{tab:row2107}). Coordinate exchange remains 10/80 overall
and 7/23 given clean correctness. The stronger ablation effect therefore
comes with reduced recovery on three families; it does not establish arbitrary
semantic steering or superior lens quality.

\iftrue\fi
\FloatBarrier

\subsection{Exchange algebra, dose and replay checks}
\label{app:exchange-numerics}

For independent unit columns $a,b$, the sum and difference directions are
orthogonal. Coordinate exchange preserves the sum component and negates the
difference component. Writing $d=b-a$ gives
\begin{equation}
    x_{X_{\rm unit}}=x-2d\frac{d^\top x}{d^\top d}.
\end{equation}
Consequently the fixed-activation contrast changes by $-2d^\top x$, whose
sign depends on the current activation. This is a reflection, not a rule
that always moves towards the target. With unequal raw column norms this
Euclidean reflection identity does not generally hold. The $G$ edit instead
has the non-negative local contrast change in Equation~\ref{eq:local-contrast};
repeated edits and downstream nonlinearities prevent interpreting that identity
as a guarantee about output logits or task success.

In the paired comparison (experiment 38), float64 inversion with atol zero
and rtol $10^{-10}$ retains rank two for every source--target pair at every
edited block. Default float32 pseudoinversion would retain only rank one on
20--38\% of Mamba-2 state pair--block combinations. Correcting that numerical
truncation still yields only 1/335 Mamba-2 state verbal successes, so it does
not explain the contrast with guarded transfer. The model and exact recurrence
run in float32; the pseudoinverse is computed in float64. The earlier
80/84-item protocol results retain their historical implementation.

The comparison freezes within-model trial manifests and token IDs, preserves
all trials and tests three matched-norm random controls for each $G/X$ arm.
Controls match each arm's own realised perturbation at each block and position,
including the combined $RS$ trajectory. They do not match doses between
operators. The bootstrap uses 100,000 paired cluster resamples within category:
verbal trials cluster by prompt, two-hop by item and flexible generalisation
by category/template/argument. Simultaneous percentile intervals use the
original Bonferroni level $1-0.05/78$ for the full 78-contrast family. This
family tests $G-X$, not $G-N$, joint-minus-residual effects or post-hoc
output categories.

\paragraph{Mamba-3 reproducibility qualification.}
The historical signed-clamp replay agrees on every suite success. Verbal
replay differs on one of 728 trials: \texttt{organ-11-t02} changes from target
rank five to six under $N:RS$, giving 703 rather than 704 top-five successes.
The kernel-pinning investigation measured context-dependent logit differences
of 0.20--0.22, including an argmax change. We retain the final pinned-context
measurements and all trials, including $X_{\rm unit}:RS=207/728$; archived
\texttt{valid\_gate: false} flags preserve the failed strict historical replay
criterion. Thus the 26 verbal contrasts, including six Mamba-3 contrasts,
must not be described as all passing that gate. Mamba-3 $G:R$ verbal and
$G:S$ flex-gen measurements were also remeasured after one-success reuse
discrepancies (712 to 713 and 9 to 10 respectively). Both records remain
archived. Legacy exchange candidates with different source/target token IDs
were remeasured rather than reused even when success counts agreed.

\paragraph{Direction and budget controls on Mamba-2.}
The separate frozen state-sign study (experiment 30) replays the 335 verbal
trials. Guarded \jfull{} and \jread{} state edits give 258 and 188 top-five
successes. Reversing the guarded displacement gives zero for both, with target
rank worsening on 335 and 331 trials. A fixed relative per-site budget of
$2.6035\times10^{-4}$, calibrated on development prompts, gives 49 and 47
successes and improves target rank on 332/335 trials for each lens, with none
worsened. All three random seeds give zero. These controls support direction
dependence at a clamp-typical budget; they do not establish a dose-matched
$G-X$ advantage. Persistent guarded edits follow different trajectories and
can accumulate larger realised displacements. Raw residual and state norms
are also not comparable units of dose.

\iftrue
\section{Corrected paired verbal-report experiment}
\label{app:verbal-report-details}

\paragraph{What the experiment asks.}
Here ``verbal report'' means the model's immediate answer to a category
question, rather than a report about its own internal state. Each trial starts
from a clean prompt whose next token is a valid category member (the
\emph{source}), selects a different category member that the clean model ranks
outside its top ten (the \emph{target}), and edits the corresponding concept
directions using $N$, $G$ or $X$. We then ask whether the target enters the model's
next-token top five. Residual, recurrent-state and simultaneous interventions
use the same eligible source--target trial within each model; matched-norm
random directions provide controls.

\paragraph{Worked example.}
One frozen Mamba-1 trial uses the following prompt:

\begin{quote}
\small\ttfamily
Q: Name a fruit.\\
A: Grape\\[2pt]
Q: Name an instrument.\\
A: Drum\\[2pt]
Q: Name a bird.\\
A: Finch\\[2pt]
Q: Name a country.\\
A:
\end{quote}
The clean answer is \emph{Canada}; the selected target is \emph{Brazil}, which
is only rank 20 before intervention. Table~\ref{tab:verbal-worked-example}
retains the historical signed-clamp outcome. The residual clamp raises \emph{Brazil} to rank 1,
whereas the state clamp alone leaves it at rank 16. Applying both clamps also
places \emph{Brazil} first, but adds nothing beyond the residual clamp in this
trial. The matched random interventions leave the clean answer at the top.

\begin{table}[H]
\centering
\small
\caption{Historical signed-clamp ($N$) outcome on Mamba-1 trial
\texttt{country-05-t00}; the random rows are its original matched controls.}
\label{tab:verbal-worked-example}
\begin{tabular}{lrrc}
\toprule
Condition & Target rank & Top-1 word & Top-5 success \\
\midrule
Clean & 20 & Canada & No \\
Residual clamp & 1 & Brazil & Yes \\
State clamp & 16 & France & No \\
Simultaneous clamps & 1 & Brazil & Yes \\
Matched random residual & 23 & Canada & No \\
Matched random state & 20 & Canada & No \\
Matched random simultaneous & 22 & Canada & No \\
\bottomrule
\end{tabular}
\end{table}

The final paired comparison adds the following outcomes on this same trial;
it does not select a new favourable example. Unit exchange leaves Brazil at
ranks 16, 11 and 14 for $R/S/RS$, with \texttt{\char32 France} top-ranked in
all three. Guarded transfer gives ranks 1, 2 and 11,560: $G:R$ outputs
\texttt{\char32 Brazil}, $G:S$ outputs \texttt{\char32 France}, and $G:RS$
outputs the token fragment \texttt{\char32 b}. Thus the state guarded edit
succeeds at top five while joint guarding fails, despite a large target-minus-
source contrast. Neither that contrast nor a successful residual edit guarantees
the target is the most competitive output under a joint intervention.

\paragraph{From prompts to eligible trials.}
For each model, format choice uses five development variants in each of 14
categories; the remaining ten variants per category give 140 evaluation
prompts. A clean prompt is counted as valid only when its greedy continuation
starts a complete single-token category member (or frozen alias), with a
leading-space alphabetic token, and does not repeat a demonstration answer.
Thus \emph{valid prompts/140} is a clean-answer validity check, not conventional
accuracy: an open question such as ``Name a country'' has many acceptable
answers. Token fragments such as ``Viol'' or ``New'', and non-category outputs,
are invalid.

For every valid prompt, the evaluator considers up to ten targets in frozen
order. A target is eligible only when it is exactly one additional token at the
answer boundary, leaves prefix tokenisation unchanged, and has clean rank
greater than ten. One valid prompt can therefore yield several source--target
trials. Table~\ref{tab:verbal-denominators} should be read left to right as a
funnel: its final column is neither a second prompt count nor a number to add to
the preceding column.

\begin{table}[htbp]
\centering
\small
\caption{Corrected verbal-report denominator funnel.}
\label{tab:verbal-denominators}
\begin{tabular}{lrrr}
\toprule
Model & Evaluation prompts & Valid clean prompts & Eligible target swaps \\
\midrule
Mamba-1 2.8B & 140 & 110 & 446 \\
Mamba-2 2.7B & 140 & 99 & 335 \\
Mamba-3 SISO 1.5B & 140 & 124 & 728 \\
Nemotron-3.5 hybrid & 140 & 134 & 621 \\
Pythia-2.8B & 140 & 107 & 399 \\
\bottomrule
\end{tabular}
\end{table}

Consequently, 446/446 means that the target entered Mamba-1's top five on all
446 eligible source--target trials under the residual clamp; it does not mean
446 prompts or 446 correct clean answers. Conditions are paired over the same
eligible trials within a model, but the trial sets are not identical across
models. Counts from residual and state conditions must therefore be compared,
not added: all 32 Mamba-1 signed-clamp state successes, for example, are
already among its 446 signed-clamp residual successes. This overlap result
describes $N$; it is not a claim that state adds no successes under $G$ or $X$.

\subsection{Full operator comparison and controls}

The main-text comparison uses simultaneous residual--state edits for recurrent
models and residual edits for Pythia, matching Table~\ref{tab:workspace-package}.
Separate state-only guarded edits place the target in the top five on
290/446 Mamba-1, 258/335 Mamba-2, 662/728 Mamba-3 and 489/621 Nemotron
trials. Unit-vector state exchange gives 31, 1, 7 and 0 successes respectively.
The complete tables below retain the historical signed clamp, both exchange
conventions and every measured intervention location.

All 26 paired verbal $G-X$ contrasts have positive simultaneous intervals in
the frozen 78-contrast family, including the six Mamba-3 contrasts retained
with the kernel-sensitivity qualification in Appendix~\ref{app:exchange-numerics}.
The contrasts include both exchange conventions and are not independent
replications. Three matched-norm random seeds give 0/0/0 state successes
for Mamba-1 and Mamba-2, 3/5/10 for Mamba-3 and 17/11/3 for Nemotron,
well below $G:S$. Mamba-2's 203/335 top-one successes also show that its
effect is not solely a top-five threshold artefact. The tables below give the complete counts, controls, top-one sensitivity
and simultaneous intervals.

Joint $G:RS$ exceeds $G:R$ by 9 Mamba-1, 42 Mamba-2 and 16 Nemotron
successes, but falls by 36 on Mamba-3. These are descriptive differences
on paired trials, not unions of separate runs or confirmatory tests of an
additional state effect. In particular, overlap claims from the historical
signed clamp apply to $N$, not to every intervention. Native doses differ
between $G$ and $X$; the comparison establishes higher target-report success
in these settings, not greater efficiency at equal dose or an unmeasured
increase in average target probability.

\begin{table}[htbp]
\centering\small
\setlength{\tabcolsep}{4pt}
\caption{Verbal report: all operators and interfaces. Counts use the stated $n$; C is clean success under the swapped-target criterion, not original-answer competence. The final three columns list matched-norm control successes for seeds 0/1/2 separately. U and Rw denote unit and raw exchange. Mamba-3 verbal includes the replay-qualified measurements.}
\label{tab:operators-verbal}
\begin{tabular}{llrrrrrrrrr}
\toprule
Model & Site & $n$ & C & $N$ & $G$ & U & Rw & $G$ ctrl & U ctrl & Rw ctrl \\
\midrule
Mamba-1 & R & 446 & 0 & 446 & 417 & 87 & 90 & 5/4/3 & 2/1/0 & 2/1/0 \\
Mamba-1 & S & 446 & 0 & 32 & 290 & 31 & 27 & 0/0/0 & 0/0/0 & 0/0/0 \\
Mamba-1 & RS & 446 & 0 & 446 & 426 & 85 & 87 & 7/6/5 & 2/0/1 & 2/0/1 \\
Mamba-2 & R & 335 & 0 & 44 & 291 & 120 & 121 & 0/0/0 & 0/1/1 & 3/2/4 \\
Mamba-2 & S & 335 & 0 & 0 & 258 & 1 & 1 & 0/0/0 & 0/0/0 & 0/0/0 \\
Mamba-2 & RS & 335 & 0 & 47 & 333 & 114 & 117 & 1/4/0 & 0/1/0 & 4/2/5 \\
Mamba-3 SISO & R & 728 & 0 & 700 & 713 & 209 & 201 & 13/20/18 & 2/1/1 & 2/1/1 \\
Mamba-3 SISO & S & 728 & 0 & 115 & 662 & 7 & 7 & 3/5/10 & 0/0/0 & 0/0/0 \\
Mamba-3 SISO & RS & 728 & 0 & 703 & 677 & 207 & 198 & 0/0/0 & 2/0/1 & 2/0/1 \\
Nemotron & R & 621 & 0 & 587 & 605 & 121 & 126 & 4/6/8 & 6/6/6 & 7/6/4 \\
Nemotron & S & 621 & 0 & 7 & 489 & 0 & 0 & 17/11/3 & 0/0/0 & 0/0/0 \\
Nemotron & RS & 621 & 0 & 583 & 621 & 130 & 136 & 0/0/0 & 4/6/6 & 7/5/5 \\
Pythia & R & 399 & 0 & 399 & 298 & 242 & 228 & 2/5/4 & 3/0/1 & 3/0/1 \\
\bottomrule
\end{tabular}
\end{table}

\begin{table}[htbp]
\centering\small
\setlength{\tabcolsep}{4pt}
\caption{Verbal report: paired $G-X$ success-rate differences and simultaneous intervals. Both exchange conventions retain the original 78-contrast family correction; these are not independent replications.}
\label{tab:contrasts-verbal}
\begin{tabular}{llrrrr}
\toprule
Model & Site & $G-X_{\rm unit}$ & Interval & $G-X_{\rm raw}$ & Interval \\
\midrule
Mamba-1 & R & +0.740 & [+0.628, +0.844] & +0.733 & [+0.620, +0.835] \\
Mamba-1 & S & +0.581 & [+0.499, +0.654] & +0.590 & [+0.510, +0.662] \\
Mamba-1 & RS & +0.765 & [+0.672, +0.857] & +0.760 & [+0.674, +0.847] \\
Mamba-2 & R & +0.510 & [+0.402, +0.617] & +0.507 & [+0.392, +0.616] \\
Mamba-2 & S & +0.767 & [+0.704, +0.829] & +0.767 & [+0.704, +0.831] \\
Mamba-2 & RS & +0.654 & [+0.547, +0.757] & +0.645 & [+0.533, +0.750] \\
Mamba-3 SISO & R & +0.692 & [+0.622, +0.766] & +0.703 & [+0.630, +0.777] \\
Mamba-3 SISO & S & +0.900 & [+0.851, +0.942] & +0.900 & [+0.853, +0.941] \\
Mamba-3 SISO & RS & +0.646 & [+0.571, +0.718] & +0.658 & [+0.588, +0.729] \\
Nemotron & R & +0.779 & [+0.707, +0.847] & +0.771 & [+0.701, +0.838] \\
Nemotron & S & +0.787 & [+0.739, +0.840] & +0.787 & [+0.738, +0.840] \\
Nemotron & RS & +0.791 & [+0.721, +0.859] & +0.781 & [+0.713, +0.848] \\
Pythia & R & +0.140 & [+0.013, +0.272] & +0.175 & [+0.033, +0.311] \\
\bottomrule
\end{tabular}
\end{table}

\begin{table}[htbp]
\centering\small
\setlength{\tabcolsep}{4pt}
\caption{Descriptive top-one sensitivity on the same verbal-report trials. The primary endpoint remains top five.}
\label{tab:verbal-top1}
\begin{tabular}{llrrrrr}
\toprule
Model & Site & $n$ & $N$ & $G$ & $X_{\rm unit}$ & $X_{\rm raw}$ \\
\midrule
Mamba-1 & R & 446 & 436 & 376 & 34 & 28 \\
Mamba-1 & S & 446 & 2 & 202 & 1 & 3 \\
Mamba-1 & RS & 446 & 442 & 401 & 19 & 17 \\
Mamba-2 & R & 335 & 26 & 258 & 74 & 77 \\
Mamba-2 & S & 335 & 0 & 203 & 0 & 0 \\
Mamba-2 & RS & 335 & 27 & 314 & 69 & 74 \\
Mamba-3 SISO & R & 728 & 671 & 692 & 113 & 106 \\
Mamba-3 SISO & S & 728 & 35 & 590 & 0 & 0 \\
Mamba-3 SISO & RS & 728 & 675 & 674 & 94 & 86 \\
Nemotron & R & 621 & 506 & 571 & 43 & 41 \\
Nemotron & S & 621 & 0 & 282 & 0 & 0 \\
Nemotron & RS & 621 & 504 & 588 & 43 & 42 \\
Pythia & R & 399 & 399 & 252 & 122 & 124 \\
\bottomrule
\end{tabular}
\end{table}

\FloatBarrier

\section{Workspace-suite comparison conventions}
\label{app:workspace-comparison}

In Table~\ref{tab:workspace-package}, introspection uses \jread{} for all
state entries. Its injection magnitude is eight times the mean baseline
activation norm over the injection span at each edited block; this is not
eight standard deviations. Other state
readouts use \jread{} for Mamba-1/3 and Nemotron, and \jfull{} for Mamba-2.
Our first-three-row measurements use unit-vector exchange with \jfull{} state lenses;
other rows retain their own protocols. Mamba-1's suite runs use 12 sampled
state blocks with exact scans at 23 source blocks, while its residual edits
cover all 33 band blocks. Its verbal-report run instead uses all 33 state-band
blocks with scans at 63 source blocks. Mamba-2 scans all 63 and edits all 33
band blocks; Mamba-3 scans 23 and edits all 13 band blocks. For rank capacity we
bound the desired full-band residual--\jfull{}-state union because the saved
rank-based files retain counts but not held-word identities: the union is at
least the state count and at most the sum of the residual and state counts.

The block package comprises the residual stream and recurrent state wherever
one exists; Pythia, like Qwen, has only the residual interface. Nemotron has
52 blocks. Residual edits cover all 27 blocks in its intervention band
13--39, while state edits cover the 12 Mamba blocks in that band; the read-only
state suite reads all 23 Mamba blocks. Thus its residual and state edits are
not matched in block coverage. Pythia uses the fixed protocol band 8--24.
All lenses follow the 1000-prompt fitting recipe.
Nemotron and Pythia summoning entries are exact zeroes in all recorded fields.

Under the historical signed clamp $N$, Nemotron's simultaneous edits yield 31/90 two-hop swaps and 78/192
flexible-generalisation redirects, compared with 31/90 and 77/192 for the
residual clamp alone and zero for the \jfull{} state clamp alone.
The refitted residual lens reproduces both residual-only counts from the
original conforming fit; we retain that fit's remaining residual-suite results.
Pythia's 9/90 is the workspace clamp-transfer probe swap (clean baseline
19/90), distinct from its 3/80 coordinate-exchange result under the recovery
protocol. Nemotron's high state introspection therefore coexists with little
additional steering under these simultaneous clamps.

\paragraph{Verbal-report protocols.}
Nemotron's historical signed \jfull{} clamp reaches 583/621 targets, slightly
fewer than the residual clamp's 587/621 (584/621 with simultaneous \jread{}
edits). Under that signed rule, neither state lens yields additional state-only
successes: $|S\setminus R|=0$ in both cases. Pythia's residual clamp reaches 399/399
targets. The eligible-trial denominators are summarised in
Table~\ref{tab:verbal-denominators}.

Qwen's historical signed-clamp results are 85/124 verbal targets (69\%),
47/90 two-hop answers and 80/192 flexible-generalisation answers. The verbal
and flex-gen reproduction explicitly subtracts the signed source projection
and adds it along the target direction at unit strength. These values are not
exchange measurements: Table~\ref{tab:workspace-package} quotes them from
\citet{wang2026looped} as marked reference values, not reproduced in this study.
The verbal run uses fourteen category prompts, with up to ten
targets per prompt initially outside the output top ten. All logged clean
answers are valid category members; however, the evaluator does not enforce
clean-answer validity or whole-word single-token targets, using each candidate's
first token. The corrected evaluation explicitly checks both, yielding
446/335/728 eligible trials for Mamba-1/2/3, 621 for Nemotron and 399 for
Pythia, from 140 prompts per model (ten evaluation variants in each of
14 categories, after setting aside five development variants)
(Appendix~\ref{app:verbal-report-details}). Trials are paired across edits
within each model, not across models. Thus 124 counts eligible target swaps,
not prompts or clean correct answers. The larger denominators here primarily
reflect the tenfold increase in evaluation prompts; they also depend on clean
answer validity, tokenisation and the target-rank filter. Both protocols score the target reaching
the next-token top five, but use different prompts and eligibility rules.

\newpage
\subsection{Worked introspection example}
\label{app:introspection-example}

The initial instruction tells the model that a researcher may inject a word
concept into its activations, and asks it to detect and identify the concept.
The first saved trial injects \texttt{lightning}, which is absent from the
prompt text. Below are the exact final question and supplied answer prefix;
chat-template markers are omitted.
\begin{quote}
\small
\textbf{User:} Trial 1: Do you detect an injected thought? If so, what is
the thought about?

\textbf{Assistant prefix (supplied):} Yes, I detect an injected thought.
The thought is about the word "

\textbf{Top predicted next token:} \texttt{lightning}
\end{quote}
With a persistent \jread{} state injection at eight times the mean baseline
activation norm during the question,
both Mamba-3 and Nemotron rank \texttt{lightning} first. Without injection,
its ranks are 18,967 and 25,338 respectively. The recorded answer is this
single next-token prediction, not a freely generated sentence: the affirmative
prefix is supplied by the evaluator. Thus the example shows that the model
can identify externally inserted content under a reporting prompt; it does
not establish spontaneous detection of an intervention or reliable reporting
of its unedited state. Across all targets, the corresponding top-1 counts are
49/101 and 93/99 (Table~\ref{tab:workspace-package}).

\subsection{Worked retained-word example}
\label{app:capacity-example}

The first Mamba-3 trial uses the following complete input, with 20 words from
each of four families (countries, first names, surnames and cities):
\begin{quote}
\small
Remember this list: Jamaica, Venezuela, Canada, Singapore, Norway, Somalia,
Laos, Iraq, Denmark, Bulgaria, Ireland, Bolivia, Slovakia, Thailand, Portugal,
Slovenia, Finland, Indonesia, Turkey, China, Marvin, Craig, Mike, Felix, Jim,
Dale, Lee, Matt, Ray, Simon, John, Barry, Terry, Gavin, Eric, Randy, Oliver, Roy,
Noah, Brandon, Young, Cook, Wright, Smith, Bailey, Brown, Fox, Edwards, Morgan,
Graham, Ferguson, Richardson, Tucker, Wood, Baker, Watson, Knight, Campbell,
Price, Burke, Lagos, Atlanta, Lima, Osaka, Ankara, Beijing, Frankfurt, Orlando,
Beirut, Bangkok, London, Moscow, Dublin, Istanbul, Amsterdam, Kiev, Boston,
Vancouver, Tokyo, Shanghai
\end{quote}
There is no subsequent recall question or generated answer. Instead, at each
list boundary, we ask of the \jfull{} state lens: how many words already read
appear among its top-five predictions at at least one evaluated block?
After 20, 40, 60 and 80 words, the recorded answers for this list are
\textbf{6, 6, 5 and 9 words}, respectively. A count can exceed five because
different blocks can recover different words.

For this trial, the mean count over positions 61--80 is 6.3; averaging the
same final-quarter counts over all ten lists gives 5.92, rounded to the
5.9-word state lower bound in Table~\ref{tab:workspace-package}.
This measures how much previously presented content remains lens-readable
while new words arrive. It does not measure how many words the model can
recite: the saved rank-based results contain counts, not the identities of
the recovered words or a verbal recall response.
\FloatBarrier
\newpage
\fi

\iftrue
\section{Paired task outcomes and saved-output diagnostics}
\label{app:twohop-output}

The final paired suite comparison uses the full 90-item two-hop cohort and
192 flexible-generalisation swaps per model. The latter reuse 64 prompts
three times; they are not 192 independent baseline observations. The full
operator/control tables preserve these denominators, and the interval tables
retain the frozen 78-family correction.
Across the 52 frozen suite $G-X$ contrasts, four favour $G$, seven favour
$X$ and 41 are inconclusive at the simultaneous level.

\begin{table}[htbp]
\centering\small
\setlength{\tabcolsep}{4pt}
\caption{Two-hop reasoning: all operators and interfaces. Counts use the stated $n$; C is clean success under the swapped-target criterion, not original-answer competence. The final three columns list matched-norm control successes for seeds 0/1/2 separately. U and Rw denote unit and raw exchange. Mamba-3 verbal includes the replay-qualified measurements.}
\label{tab:operators-probe_swap}
\begin{tabular}{llrrrrrrrrr}
\toprule
Model & Site & $n$ & C & $N$ & $G$ & U & Rw & $G$ ctrl & U ctrl & Rw ctrl \\
\midrule
Mamba-1 & R & 90 & 1 & 10 & 5 & 7 & 7 & 1/1/3 & 1/1/1 & 1/1/1 \\
Mamba-1 & S & 90 & 1 & 1 & 1 & 1 & 1 & 1/1/1 & 1/1/1 & 1/1/1 \\
Mamba-1 & RS & 90 & 1 & 11 & 4 & 6 & 6 & 1/2/3 & 1/1/1 & 1/1/1 \\
Mamba-2 & R & 90 & 1 & 1 & 12 & 7 & 7 & 1/1/1 & 1/2/1 & 0/2/1 \\
Mamba-2 & S & 90 & 1 & 1 & 2 & 1 & 1 & 1/1/1 & 1/1/1 & 1/1/1 \\
Mamba-2 & RS & 90 & 1 & 1 & 6 & 7 & 5 & 1/1/1 & 1/2/1 & 0/3/1 \\
Mamba-3 SISO & R & 90 & 1 & 3 & 0 & 6 & 6 & 0/0/0 & 1/1/1 & 1/1/1 \\
Mamba-3 SISO & S & 90 & 1 & 12 & 16 & 1 & 1 & 1/1/1 & 1/1/1 & 1/1/1 \\
Mamba-3 SISO & RS & 90 & 1 & 14 & 0 & 6 & 6 & 0/0/0 & 1/1/1 & 1/1/1 \\
Nemotron & R & 90 & 0 & 31 & 16 & 16 & 18 & 0/0/0 & 0/0/0 & 0/0/0 \\
Nemotron & S & 90 & 0 & 0 & 5 & 0 & 0 & 2/0/0 & 0/0/0 & 0/0/0 \\
Nemotron & RS & 90 & 0 & 31 & 1 & 15 & 17 & 0/0/0 & 0/0/0 & 0/0/0 \\
Pythia & R & 90 & 2 & 9 & 2 & 5 & 5 & 0/0/1 & 2/2/2 & 2/2/2 \\
\bottomrule
\end{tabular}
\end{table}

\begin{table}[htbp]
\centering\small
\setlength{\tabcolsep}{4pt}
\caption{Two-hop reasoning: paired $G-X$ success-rate differences and simultaneous intervals. Both exchange conventions retain the original 78-contrast family correction; these are not independent replications.}
\label{tab:contrasts-probe_swap}
\begin{tabular}{llrrrr}
\toprule
Model & Site & $G-X_{\rm unit}$ & Interval & $G-X_{\rm raw}$ & Interval \\
\midrule
Mamba-1 & R & -0.022 & [-0.089, +0.056] & -0.022 & [-0.089, +0.056] \\
Mamba-1 & S & +0.000 & [+0.000, +0.000] & +0.000 & [+0.000, +0.000] \\
Mamba-1 & RS & -0.022 & [-0.089, +0.044] & -0.022 & [-0.089, +0.044] \\
Mamba-2 & R & +0.056 & [-0.033, +0.144] & +0.056 & [-0.011, +0.133] \\
Mamba-2 & S & +0.011 & [+0.000, +0.044] & +0.011 & [+0.000, +0.044] \\
Mamba-2 & RS & -0.011 & [-0.100, +0.067] & +0.011 & [-0.067, +0.078] \\
Mamba-3 SISO & R & -0.067 & [-0.122, -0.022] & -0.067 & [-0.122, -0.022] \\
Mamba-3 SISO & S & +0.167 & [+0.089, +0.233] & +0.167 & [+0.089, +0.233] \\
Mamba-3 SISO & RS & -0.067 & [-0.122, -0.022] & -0.067 & [-0.122, -0.022] \\
Nemotron & R & +0.000 & [-0.067, +0.078] & -0.022 & [-0.111, +0.056] \\
Nemotron & S & +0.056 & [+0.000, +0.122] & +0.056 & [+0.000, +0.122] \\
Nemotron & RS & -0.156 & [-0.200, -0.089] & -0.178 & [-0.233, -0.111] \\
Pythia & R & -0.033 & [-0.100, +0.022] & -0.033 & [-0.100, +0.022] \\
\bottomrule
\end{tabular}
\end{table}

\begin{table}[htbp]
\centering\small
\setlength{\tabcolsep}{4pt}
\caption{Flexible generalisation: all operators and interfaces. Counts use the stated $n$; C is clean success under the swapped-target criterion, not original-answer competence. The final three columns list matched-norm control successes for seeds 0/1/2 separately. U and Rw denote unit and raw exchange. Mamba-3 verbal includes the replay-qualified measurements.}
\label{tab:operators-flexgen}
\begin{tabular}{llrrrrrrrrr}
\toprule
Model & Site & $n$ & C & $N$ & $G$ & U & Rw & $G$ ctrl & U ctrl & Rw ctrl \\
\midrule
Mamba-1 & R & 192 & 4 & 37 & 18 & 32 & 33 & 2/2/3 & 6/4/3 & 6/4/3 \\
Mamba-1 & S & 192 & 4 & 3 & 3 & 3 & 3 & 4/4/4 & 4/4/3 & 4/4/3 \\
Mamba-1 & RS & 192 & 4 & 37 & 18 & 30 & 31 & 2/2/3 & 6/4/3 & 6/4/3 \\
Mamba-2 & R & 192 & 2 & 2 & 8 & 12 & 12 & 0/0/0 & 2/1/3 & 2/2/3 \\
Mamba-2 & S & 192 & 2 & 3 & 2 & 3 & 3 & 2/2/2 & 2/2/2 & 2/2/2 \\
Mamba-2 & RS & 192 & 2 & 2 & 23 & 12 & 11 & 0/0/0 & 2/1/3 & 2/2/3 \\
Mamba-3 SISO & R & 192 & 2 & 12 & 6 & 21 & 21 & 1/1/0 & 4/0/1 & 4/0/1 \\
Mamba-3 SISO & S & 192 & 2 & 10 & 10 & 2 & 2 & 2/2/2 & 2/2/2 & 2/2/2 \\
Mamba-3 SISO & RS & 192 & 2 & 24 & 4 & 23 & 24 & 0/0/0 & 4/0/1 & 4/0/1 \\
Nemotron & R & 192 & 0 & 77 & 29 & 32 & 35 & 1/1/0 & 1/1/0 & 1/1/0 \\
Nemotron & S & 192 & 0 & 0 & 12 & 0 & 0 & 0/0/0 & 0/0/0 & 0/0/0 \\
Nemotron & RS & 192 & 0 & 78 & 17 & 29 & 29 & 0/0/0 & 1/1/0 & 1/1/0 \\
Pythia & R & 192 & 3 & 42 & 5 & 13 & 12 & 1/0/2 & 3/3/3 & 3/3/3 \\
\bottomrule
\end{tabular}
\end{table}

\begin{table}[htbp]
\centering\small
\setlength{\tabcolsep}{4pt}
\caption{Flexible generalisation: paired $G-X$ success-rate differences and simultaneous intervals. Both exchange conventions retain the original 78-contrast family correction; these are not independent replications.}
\label{tab:contrasts-flexgen}
\begin{tabular}{llrrrr}
\toprule
Model & Site & $G-X_{\rm unit}$ & Interval & $G-X_{\rm raw}$ & Interval \\
\midrule
Mamba-1 & R & -0.073 & [-0.188, +0.047] & -0.078 & [-0.193, +0.042] \\
Mamba-1 & S & +0.000 & [+0.000, +0.000] & +0.000 & [+0.000, +0.000] \\
Mamba-1 & RS & -0.062 & [-0.172, +0.052] & -0.068 & [-0.177, +0.052] \\
Mamba-2 & R & -0.021 & [-0.120, +0.078] & -0.021 & [-0.115, +0.078] \\
Mamba-2 & S & -0.005 & [-0.026, +0.000] & -0.005 & [-0.026, +0.000] \\
Mamba-2 & RS & +0.057 & [-0.036, +0.146] & +0.062 & [-0.016, +0.141] \\
Mamba-3 SISO & R & -0.078 & [-0.172, +0.016] & -0.078 & [-0.172, +0.016] \\
Mamba-3 SISO & S & +0.042 & [-0.010, +0.120] & +0.042 & [-0.015, +0.120] \\
Mamba-3 SISO & RS & -0.099 & [-0.198, +0.000] & -0.104 & [-0.208, -0.010] \\
Nemotron & R & -0.016 & [-0.115, +0.083] & -0.031 & [-0.135, +0.073] \\
Nemotron & S & +0.062 & [+0.010, +0.125] & +0.062 & [+0.016, +0.125] \\
Nemotron & RS & -0.062 & [-0.156, +0.031] & -0.062 & [-0.156, +0.031] \\
Pythia & R & -0.042 & [-0.120, +0.041] & -0.036 & [-0.109, +0.036] \\
\bottomrule
\end{tabular}
\end{table}

\FloatBarrier

\subsection{Output classification and original-answer competence}
The audit validates all 450 prompt/trial records against their manifests and
five pinned tokenisers. It assigns categories in this order: requested-answer
success, edited intermediate, original answer, unresolved original answer,
source intermediate, other failure with the same clean first token, other
failure with a changed first token. The unresolved category permits missing
continuations but is zero here. Requested and original answers both use the
harness's leading-token criterion, with the saved second step when the first
token is whitespace; whitespace alone is not answer retention. The weaker
same-first-token flag can also hold for other categories and does not imply
unchanged logits or the same complete answer.

\begin{table}[htbp]
\centering\small
\setlength{\tabcolsep}{4pt}
\caption{Two-hop output diagnostics on 90 trials per model, at the Table~\ref{tab:workspace-package} interfaces: simultaneous residual and state edits for recurrent models, residual edits for Pythia. Exchange-failure columns give the total and the subset retaining the clean first token. Edited-intermediate columns count outputs across all 90 trials under each method. Exchange uses the unit-vector convention.}
\label{tab:twohop-failure-summary}
\begin{tabular}{lrrrr}
\toprule
& \multicolumn{2}{c}{\textbf{Exchange failures}} & \multicolumn{2}{c}{\textbf{Edited-intermediate outputs}} \\
\cmidrule(lr){2-3}\cmidrule(lr){4-5}
\textbf{Model} & \textbf{Total} & \shortstack{\textbf{Retain clean}\\\textbf{first token}} & \shortstack{\textbf{Coordinate}\\\textbf{exchange}} & \shortstack{\textbf{Sign-guarded}\\\textbf{steering}} \\
\midrule
Mamba-1 & 84 & 73 & 1 & 55 \\
Mamba-2 & 83 & 69 & 2 & 37 \\
Mamba-3 SISO & 84 & 59 & 2 & 74 \\
Nemotron & 75 & 60 & 1 & 73 \\
Pythia & 85 & 72 & 3 & 72 \\
\bottomrule
\end{tabular}
\end{table}

Table~\ref{tab:twohop-failure-summary} gives the unit-exchange failure counts
and edited-intermediate outputs shown in Figure~\ref{fig:twohop-outcomes}.
Retained tokens can be originally correct answers or existing errors,
including newline, punctuation and function-word pieces; some exchange
failures change to other outputs. All observed edited-intermediate outputs
fail the original task score and do not demonstrate internal answer substitution.

The originally correct subsets contain 20, 25, 27, 49 and 19 trials for
Mamba-1, Mamba-2, Mamba-3, Nemotron and Pythia respectively. Unit exchange
preserves the original answer on 11/20, 12/25, 13/27, 25/49 and 13/19;
$G$ instead outputs the target intermediate on 12/20, 10/25, 23/27, 39/49
and 13/19. These subsets distinguish baseline failures from loss of an
initially correct answer. Original competence does not establish competence
on the substituted relation.

Tables~\ref{tab:outcomes-full-cohort} and~\ref{tab:outcomes-clean-correct}
give exhaustive counts at the main-table interfaces. Full per-trial labels,
decoded pieces, R/S/RS outcomes and source hashes accompany the reproducible
audit. The two exchange conventions show the same broad retention pattern:
$X_{\rm raw}$ retains the clean first token on 72/84, 65/85, 59/84, 59/73
and 71/85 failed trials for Mamba-1/2/3, Nemotron and Pythia respectively.
The edited-concept diagnostic compares the saved first-token ID to the
edited-intermediate ID, not to the downstream-answer ID. Every observed
concept-token output in two-hop and flex-gen fails its task score.

\begin{table}[htbp]
\centering\small
\setlength{\tabcolsep}{4pt}
\caption{Two-hop output partition: all trials. A = requested answer, I = edited intermediate, O = original answer, S = source intermediate, C/D = other failure with the same/changed clean first token. Categories are exclusive and exhaustive; original-answer scoring uses the saved second step after whitespace where required. No original-answer cases are unresolved. RS for recurrent models, R for Pythia.}
\label{tab:outcomes-full-cohort}
\begin{tabular}{llrrrrrrr}
\toprule
Model & Edit & $n$ & A & I & O & S & C & D \\
\midrule
Mamba-1 & Xunit & 90 & 6 & 1 & 11 & 1 & 61 & 10 \\
Mamba-1 & G & 90 & 4 & 55 & 0 & 0 & 7 & 24 \\
Mamba-1 & N & 90 & 11 & 12 & 2 & 0 & 48 & 17 \\
Mamba-1 & Xraw & 90 & 6 & 1 & 12 & 1 & 59 & 11 \\
Mamba-2 & Xunit & 90 & 7 & 2 & 12 & 2 & 55 & 12 \\
Mamba-2 & G & 90 & 6 & 37 & 0 & 0 & 21 & 26 \\
Mamba-2 & N & 90 & 1 & 3 & 25 & 4 & 50 & 7 \\
Mamba-2 & Xraw & 90 & 5 & 2 & 14 & 2 & 49 & 18 \\
Mamba-3 SISO & Xunit & 90 & 6 & 2 & 14 & 5 & 43 & 20 \\
Mamba-3 SISO & G & 90 & 0 & 74 & 0 & 0 & 1 & 15 \\
Mamba-3 SISO & N & 90 & 14 & 21 & 5 & 0 & 31 & 19 \\
Mamba-3 SISO & Xraw & 90 & 6 & 2 & 14 & 5 & 43 & 20 \\
Nemotron & Xunit & 90 & 15 & 1 & 26 & 4 & 33 & 11 \\
Nemotron & G & 90 & 1 & 73 & 0 & 0 & 0 & 16 \\
Nemotron & N & 90 & 31 & 5 & 6 & 1 & 19 & 28 \\
Nemotron & Xraw & 90 & 17 & 1 & 25 & 3 & 33 & 11 \\
Pythia & Xunit & 90 & 5 & 3 & 13 & 1 & 58 & 10 \\
Pythia & G & 90 & 2 & 72 & 0 & 1 & 6 & 9 \\
Pythia & N & 90 & 9 & 34 & 1 & 0 & 41 & 5 \\
Pythia & Xraw & 90 & 5 & 3 & 13 & 2 & 57 & 10 \\
\bottomrule
\end{tabular}
\end{table}

\begin{table}[htbp]
\centering\small
\setlength{\tabcolsep}{4pt}
\caption{Two-hop output partition: originally correct trials. A = requested answer, I = edited intermediate, O = original answer, S = source intermediate, C/D = other failure with the same/changed clean first token. Categories are exclusive and exhaustive; original-answer scoring uses the saved second step after whitespace where required. No original-answer cases are unresolved. RS for recurrent models, R for Pythia.}
\label{tab:outcomes-clean-correct}
\begin{tabular}{llrrrrrrr}
\toprule
Model & Edit & $n$ & A & I & O & S & C & D \\
\midrule
Mamba-1 & Xunit & 20 & 3 & 1 & 11 & 0 & 0 & 5 \\
Mamba-1 & G & 20 & 3 & 12 & 0 & 0 & 0 & 5 \\
Mamba-1 & N & 20 & 6 & 6 & 2 & 0 & 0 & 6 \\
Mamba-1 & Xraw & 20 & 3 & 1 & 12 & 0 & 0 & 4 \\
Mamba-2 & Xunit & 25 & 5 & 1 & 12 & 0 & 0 & 7 \\
Mamba-2 & G & 25 & 4 & 10 & 0 & 0 & 0 & 11 \\
Mamba-2 & N & 25 & 0 & 1 & 23 & 0 & 0 & 1 \\
Mamba-2 & Xraw & 25 & 4 & 1 & 14 & 0 & 0 & 6 \\
Mamba-3 SISO & Xunit & 27 & 2 & 1 & 13 & 0 & 0 & 11 \\
Mamba-3 SISO & G & 27 & 0 & 23 & 0 & 0 & 0 & 4 \\
Mamba-3 SISO & N & 27 & 11 & 7 & 4 & 0 & 0 & 5 \\
Mamba-3 SISO & Xraw & 27 & 2 & 1 & 13 & 0 & 0 & 11 \\
Nemotron & Xunit & 49 & 15 & 1 & 25 & 2 & 0 & 6 \\
Nemotron & G & 49 & 1 & 39 & 0 & 0 & 0 & 9 \\
Nemotron & N & 49 & 29 & 4 & 5 & 0 & 0 & 11 \\
Nemotron & Xraw & 49 & 17 & 1 & 24 & 1 & 0 & 6 \\
Pythia & Xunit & 19 & 2 & 0 & 13 & 0 & 0 & 4 \\
Pythia & G & 19 & 1 & 13 & 0 & 0 & 0 & 5 \\
Pythia & N & 19 & 6 & 10 & 1 & 0 & 0 & 2 \\
Pythia & Xraw & 19 & 2 & 0 & 13 & 0 & 0 & 4 \\
\bottomrule
\end{tabular}
\end{table}

\begin{table}[htbp]
\centering\small
\setlength{\tabcolsep}{4pt}
\caption{Two-hop output partitions at separate residual and state interfaces (90 trials per row). Categories A/I/O/S/C/D are as in Table~\ref{tab:outcomes-full-cohort}. Together with that table, these cover every N/G/X interface; Pythia has residual only.}
\label{tab:outcomes-other}
\begin{tabular}{lllrrrrrr}
\toprule
Model & Site & Edit & A & I & O & S & C & D \\
\midrule
Mamba-1 & R & N & 10 & 12 & 2 & 0 & 48 & 18 \\
Mamba-1 & R & G & 5 & 53 & 0 & 0 & 9 & 23 \\
Mamba-1 & R & Xunit & 7 & 1 & 10 & 1 & 62 & 9 \\
Mamba-1 & R & Xraw & 7 & 1 & 10 & 1 & 64 & 7 \\
Mamba-1 & S & N & 1 & 0 & 20 & 1 & 64 & 4 \\
Mamba-1 & S & G & 1 & 0 & 20 & 1 & 64 & 4 \\
Mamba-1 & S & Xunit & 1 & 0 & 19 & 1 & 65 & 4 \\
Mamba-1 & S & Xraw & 1 & 0 & 17 & 2 & 63 & 7 \\
Mamba-2 & R & N & 1 & 3 & 26 & 4 & 50 & 6 \\
Mamba-2 & R & G & 12 & 33 & 1 & 0 & 25 & 19 \\
Mamba-2 & R & Xunit & 7 & 2 & 12 & 2 & 55 & 12 \\
Mamba-2 & R & Xraw & 7 & 2 & 13 & 2 & 48 & 18 \\
Mamba-2 & S & N & 1 & 0 & 25 & 3 & 59 & 2 \\
Mamba-2 & S & G & 2 & 10 & 16 & 1 & 44 & 17 \\
Mamba-2 & S & Xunit & 1 & 0 & 24 & 3 & 60 & 2 \\
Mamba-2 & S & Xraw & 1 & 0 & 25 & 3 & 59 & 2 \\
Mamba-3 SISO & R & N & 3 & 19 & 14 & 0 & 32 & 22 \\
Mamba-3 SISO & R & G & 0 & 63 & 0 & 0 & 3 & 24 \\
Mamba-3 SISO & R & Xunit & 6 & 3 & 13 & 5 & 42 & 21 \\
Mamba-3 SISO & R & Xraw & 6 & 3 & 13 & 4 & 43 & 21 \\
Mamba-3 SISO & S & N & 12 & 5 & 5 & 1 & 45 & 22 \\
Mamba-3 SISO & S & G & 16 & 14 & 1 & 0 & 27 & 32 \\
Mamba-3 SISO & S & Xunit & 1 & 0 & 26 & 4 & 55 & 4 \\
Mamba-3 SISO & S & Xraw & 1 & 0 & 27 & 3 & 52 & 7 \\
Nemotron & R & N & 31 & 6 & 5 & 1 & 19 & 28 \\
Nemotron & R & G & 16 & 49 & 1 & 0 & 2 & 22 \\
Nemotron & R & Xunit & 16 & 1 & 25 & 3 & 36 & 9 \\
Nemotron & R & Xraw & 18 & 1 & 23 & 3 & 35 & 10 \\
Nemotron & S & N & 0 & 0 & 42 & 2 & 32 & 14 \\
Nemotron & S & G & 5 & 31 & 2 & 0 & 3 & 49 \\
Nemotron & S & Xunit & 0 & 0 & 48 & 3 & 33 & 6 \\
Nemotron & S & Xraw & 0 & 0 & 48 & 3 & 34 & 5 \\
\bottomrule
\end{tabular}
\end{table}

\FloatBarrier

\subsection{Paired interface interactions and examples}
The Mamba-3 example in Section~\ref{sec:causal} is
\texttt{ps03-animal-nose-elephant}: elephant to snake, with tongue the
requested answer. Its $G:S$ and $G:RS$ predictions are respectively
\texttt{\char32 tongue} and \texttt{\char32 snake}. Of the 16 successful
state trials, one already meets the swapped-answer criterion under clean;
15 are new relative to clean. Eleven state successes become intermediate
outputs under joint editing. This is a paired behavioural interaction, not
evidence for a particular internal failure mechanism.
The interaction may depend on the SSM architecture, but these checkpoint
comparisons do not isolate architecture from training and other model
differences. Nor do they assign fixed reasoning or reporting roles to the
residual and state interfaces.

The transition is not exclusive to Mamba-3. Mamba-1 and Mamba-2 have 1 and 2
$G:S$ successes, all becoming intermediate outputs under $G:RS$. One in
each model already succeeds under clean, leaving 0 and 1 new state successes.
Mamba-1's \texttt{ps37-ex-planet-color-third-fourth} already predicts red;
$G:S$ retains red and $G:RS$ predicts Mars. Mamba-2's
\texttt{ps32-ex-element-state-8-26} similarly changes clean/state solid to
joint iron. Its other trial, \texttt{ps47-ex2-language-capital-Swedish},
edits Sweden to Hungary: clean predicts Stockholm, $G:S$ predicts Budapest
and $G:RS$ predicts Hungary. Their joint totals, 4/90 and 6/90, exceed
their state-only totals; paired losses on these trials do not establish
aggregate state-only superiority.

Nemotron gives 16/90 requested answers under $G:R$ but only 1/90 under
$G:RS$; eight residual successes become intermediate outputs. Its unit
exchange and guarded residual edits each give 16 successes but share only
11. Nine of its 15 successful unit joint exchanges become intermediate
outputs under guarded joint editing. These descriptive paired results show
why equal marginal totals need not identify the same successful trials.

The Einstein example uses \texttt{ps72-person-firstname-einstein}, originally
correct on all four recurrent models. Clean predicts \texttt{\char32 Albert};
unit exchange predicts that same token on Mamba-1/2/Nemotron, but
\texttt{\char32 not} on Mamba-3. Guarded joint editing predicts
\texttt{\char32 Newton} on all four; Isaac is the requested answer. These
selected illustrations are not a prevalence estimate.

State-only guarded intermediate outputs are 0/10/14/31 of 90 for
Mamba-1/2/3 and Nemotron. The frequent residual/joint intermediate-output
pattern therefore does not explain every state-only failure or show that
state cannot redirect a relation. Flexible-generalisation argument outputs
can likewise replace the function's answer---Canada rather than Ottawa---and
are rare for $G:S$ (0/1/2/4 of 192) but common for $G:RS$
(106/51/131/98); Pythia $G:R$ gives 109/192. Only the argument-output
diagnostic is audited here, not a flex-gen original-answer retention partition.

\begin{figure}[htbp]
\centering
\includegraphics[width=\linewidth]{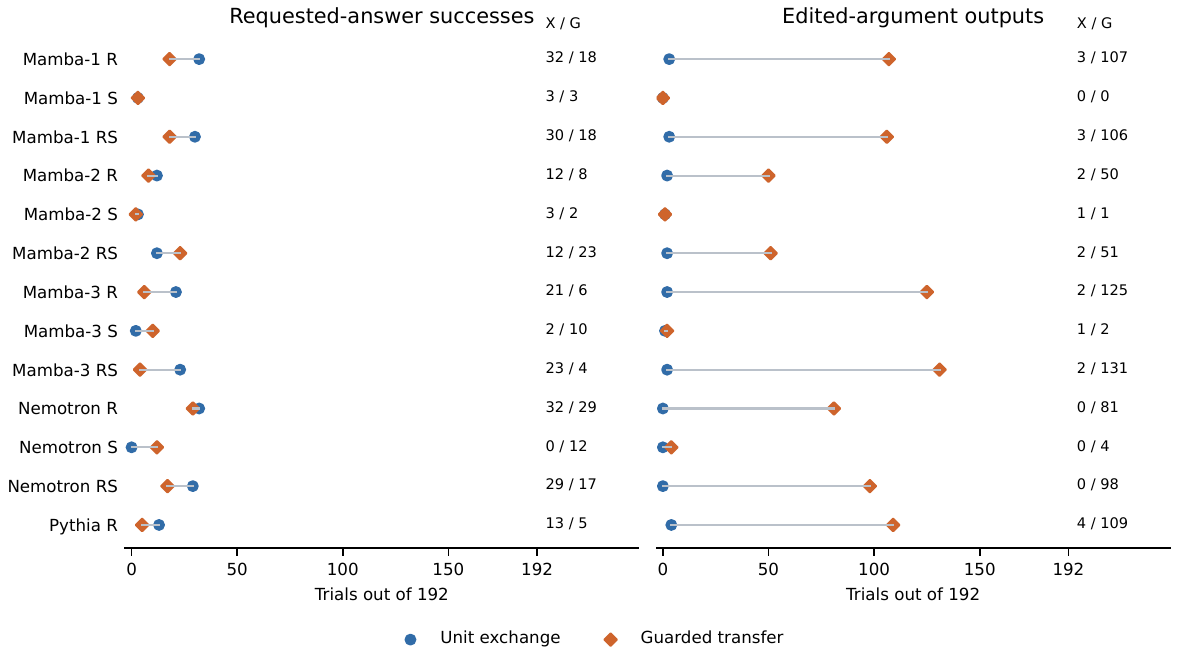}
\caption{Flexible generalisation at every measured interface. Unit exchange
and guarded transfer are compared on the same 192 swaps per model; the
right-hand counts in each panel give X/G. Requested-answer success (left)
and first-token output of the edited argument (right) are distinct endpoints.
The latter is a descriptive post-hoc diagnostic: every observed argument
output fails the task criterion. The 192 swaps reuse 64 prompts three times.}
\label{fig:flexgen-outputs}
\end{figure}
\FloatBarrier

The audit and figures are descriptive. Saved two-hop/flex-gen
\texttt{p\_new} and \texttt{rank\_new} refer to the requested answer,
not the edited concept; the latter's probability changes and complete
continuations are unavailable. Existing $G-X$ intervals test requested-answer
success, not these output categories, interface transitions or $G-N$ effects.
No new significance tests are applied to the output decomposition.
Output competition, prompt conflict, disrupted relations and accumulating
dose remain hypotheses; the saved outputs do not isolate a failure mechanism.
\fi

\section{Context dependence of Mamba-2 residual Jacobians}
\label{app:geometry}

We characterise context dependence in Mamba-2 by computing exact same-position
residual Jacobians at one random valid position in each of 200 WikiText
prompts disjoint from the fitting set, at layers $0,8,\ldots,56$. This analysis
covers the residual stream of Mamba-2, not recurrent-state Jacobians or a
comparison across Mamba generations. The mean pairwise cosine is 0.033 at
layer 0, 0.161 at layer 32, and 0.537 at layer 56. Thus local maps are strongly
context-dependent early and increasingly aligned late. The shared structure
is also asymmetric: at layer 0, a shared top-16 output subspace captures 0.663
of Jacobian energy, compared with 0.074 for the shared input subspace and
approximately 0.006 for random directions. These values describe concentration
of derivative energy, not a semantic workspace.

\iftrue
\begin{figure}[H]
\centering
\includegraphics[width=\linewidth]{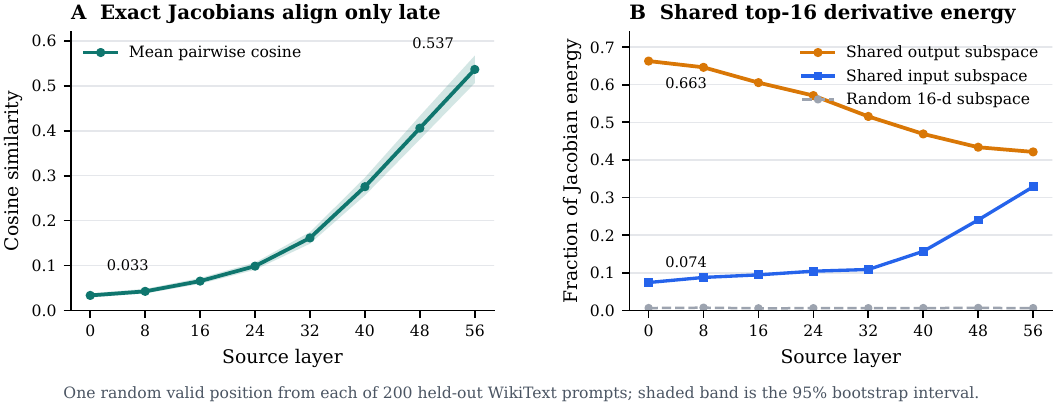}
\caption{Held-out Mamba-2 residual-Jacobian geometry across 200 WikiText
positions. Exact same-position Jacobians are weakly aligned early and become
more similar late (left). A shared top-16 output subspace captures much more
derivative energy than the corresponding input subspace, especially early
(right). These quantities diagnose context dependence; they do not explain the
intervention results.}
\label{fig:geometry}
\end{figure}
\FloatBarrier
\fi

The conforming fitted lens is not simply the small residue of these local maps.
At layers 0--24 its leading rank-one component accounts for 86\% down to 51\%
of squared Frobenius norm, yet aligns only weakly with individual held-out
Jacobians. Its norm is roughly 0.51--0.72 of a typical local Jacobian, and its
agreement with the held-out sample mean emerges only later. The origin of this
early component remains unresolved.

The norm of an empirical mean relative to a typical sample depends on sample
size even when the underlying pairwise cosine is unchanged. Consequently, an
older 32-sample mean-norm ratio is not a model invariant. All geometry values
reported here use the 200-position held-out WikiText analysis.

These observations qualify, but do not invalidate, the lens. Equation
\ref{eq:res-jacobian} averages over source positions and sums current/future
target contributions,
whereas the held-out exact maps are same-position derivatives; they are not the
same estimand. Moreover, local next-token transport is not the intermediate-
recovery metric. No estimator-rescue experiment has shown that the geometric
mismatch causes Mamba-2's clamp result, and Mamba-2's positive exchange already
precludes that simple explanation. We therefore use geometry as a diagnosis
of context dependence, not as a completed mechanism.

\section{Shared output geometry and intermediate-concept recovery}
\label{app:shared-output}

Residual and state lenses map different representations into the same final
residual coordinates. We ask whether their strongest geometric overlap also
provides a compact readout of intermediate concepts. Across Mamba-1 2.8B,
Mamba-2 2.7B and Mamba-3 SISO 1.5B, a shared rank-16 output projection retains
substantial Jacobian energy, but restricting the joint readout to it loses
much of the full readout's recovery. Removing those directions largely
preserves recovery.

\paragraph{Where the overlap is measured.}
The transported vectors $r_h=J^h_\ell h_{\ell,t}$ and
$r_s=J^s_\ell s_{\ell,t}$ in Equation~\ref{eq:joint} both have
$d_{\rm model}$ coordinates, even though the source residual and state have
different dimensions. We compare directions on this common \emph{output side}
of the lenses, before final normalisation and unembedding. By contrast,
\citet{gurnee2026verbalizable} define \jspace{} through sparse non-negative
combinations of token-labelled lens vectors in the \emph{source activation
space}. Those vectors live in the residual stream for a residual lens, and
in the recurrent-state coordinates for its state analogue. The shared output
projection studied here is therefore distinct from an intersection of the two
source-space concept dictionaries. Its usefulness for semantic readout must
be tested by decoding the projected representations.

\paragraph{Constructing the shared projection.}
We use the existing conforming lenses, with \jfull{} primary and \jread{}
as a sensitivity analysis. Separately at every source block of each model,
we obtain each lens's leading 64 left singular vectors from its output Gram
$JJ^\top$. Principal-angle analysis pairs directions $a_i,b_i$ from these
two subspaces, ordered by decreasing non-negative alignment
$c_i=a_i^\top b_i$. Their normalised midpoints define an orthogonal projector:
\begin{equation}
    q_i=\frac{a_i+b_i}{\sqrt{2+2c_i}},
    \qquad P_{\ell,k}=\sum_{i=1}^{k}q_iq_i^\top.
    \label{eq:shared-output-projector}
\end{equation}
We predeclare $k=16$ as primary and $k=64$ as sensitivity, retain projectors
at all 63/63/23 source blocks, and use no evaluation labels to select them.
Thus the primary construction takes the 16 most aligned pairs within the
two leading 64-dimensional subspaces. Controls use each lens's own leading
$k$ output directions and ten seeded Haar-random rank-$k$ projectors.
For any lens or local map $J$, retained energy is
\begin{equation}
    E(P,J)=\frac{\lVert PJ\rVert_F^2}{\lVert J\rVert_F^2}.
    \label{eq:shared-output-energy}
\end{equation}
This measures concentration of derivative sensitivity; it does not measure
the number of concepts encoded or recovered.

\paragraph{Shared sensitivity across contexts and delays.}
Table~\ref{tab:shared-output-geometry} shows strong alignment of the selected
directions and substantial energy retention in both fitted lenses. The
selected-pair alignment is higher than the direct overlap between the two
own-top-16 subspaces, because it searches within their leading 64 directions.

\begin{table}[htbp]
\centering\small
\caption{Output geometry of the fitted residual and \jfull{} state lenses,
averaged over the protocol band. Alignment is the mean squared cosine of the
16 closest principal pairs between the leading 64-dimensional subspaces.
Own overlap is the direct top-16 subspace overlap, normalised by 16.
Energy columns use the shared rank-16 projector; the random column gives
residual-lens energy under rank-16 random projectors.}
\label{tab:shared-output-geometry}
\begin{tabular}{lrrrrr}
\toprule
Model & Alignment & Own overlap & Residual energy & State energy & Random \\
\midrule
Mamba-1 & 0.887 & 0.500 & 0.381 & 0.752 & 0.0062 \\
Mamba-2 & 0.882 & 0.504 & 0.554 & 0.659 & 0.0063 \\
Mamba-3 SISO & 0.894 & 0.593 & 0.400 & 0.619 & 0.0079 \\
\bottomrule
\end{tabular}
\end{table}

On 200 held-out WikiText prompts, we also compare local residual and full-path
state derivatives from the same source position to the same final-block target
position. We probe four approximately matched depths, at blocks 8/24/40/56
for Mamba-1/2 and 3/9/15/21 for Mamba-3, with lags 0/4/16/64 tokens. Target
character boundaries are matched across tokenisers; valid-prompt counts are
200/187/175/97 for Mamba-1/2 and 192/185/170/98 for Mamba-3. Shared-16 captures
12--73\% of local-map energy across the measured cells, compared with
0.6--0.8\% for random projections. A common component coexists with context
dependence: same-context residual/state top-16 overlap at lag zero is
0.34--0.80, while cross-context overlap is 0.10--0.31. For example, Mamba-1
block 8 gives 0.802 within context and 0.119 across contexts. These quantities
do not establish a context-invariant semantic format.

\paragraph{Concept recovery within and outside the shared projection.}
We apply the fixed projectors to clean transported readouts and evaluate all
six official task families. Table~\ref{tab:shared-output-recovery} uses the
post-update, normalised joint readout: each component is projected, normalised
to unit norm, and then summed before final normalisation and unembedding.
The complement replaces $P_{\ell,16}$ with $I-P_{\ell,16}$. These are derived
readouts of the existing lenses, with unchanged model activations.

\begin{table}[htbp]
\centering\small
\caption{Normalised joint intermediate-recovery \auc{} over all source
blocks, using post-update residual and \jfull{} state readouts. Shared and
own retain 16 directions; own projects each component onto its own lens's
top-16 subspace. Random averages ten rank-16 projectors. Complement removes
the shared directions and retains $d_{\rm model}-16$ dimensions; its
comparison with full tests removal, rather than equal-rank compression.}
\label{tab:shared-output-recovery}
\begin{tabular}{llrrrrr}
\toprule
Model & Task & Full & Shared & Complement & Own & Random \\
\midrule
Mamba-1 & Multihop      & 0.554 & 0.204 & 0.534 & 0.211 & 0.019 \\
        & Multilingual & 0.710 & 0.106 & 0.666 & 0.100 & 0.042 \\
        & Order of ops.& 0.801 & 0.389 & 0.807 & 0.722 & 0.106 \\
        & Poetry       & 0.115 & 0.039 & 0.113 & 0.043 & 0.058 \\
        & Association  & 0.337 & 0.116 & 0.274 & 0.073 & 0.038 \\
        & Typo         & 0.712 & 0.150 & 0.853 & 0.079 & 0.066 \\
\midrule
Mamba-2 & Multihop      & 0.679 & 0.233 & 0.643 & 0.374 & 0.045 \\
        & Multilingual & 0.647 & 0.166 & 0.640 & 0.271 & 0.070 \\
        & Order of ops.& 0.807 & 0.428 & 0.829 & 0.690 & 0.265 \\
        & Poetry       & 0.114 & 0.040 & 0.106 & 0.059 & 0.082 \\
        & Association  & 0.231 & 0.137 & 0.195 & 0.136 & 0.049 \\
        & Typo         & 0.714 & 0.152 & 0.782 & 0.132 & 0.088 \\
\midrule
Mamba-3 SISO & Multihop      & 0.484 & 0.094 & 0.514 & 0.120 & 0.025 \\
             & Multilingual & 0.393 & 0.156 & 0.366 & 0.155 & 0.028 \\
             & Order of ops.& 0.518 & 0.367 & 0.484 & 0.432 & 0.119 \\
             & Poetry       & 0.050 & 0.003 & 0.046 & 0.003 & 0.029 \\
             & Association  & 0.141 & 0.037 & 0.115 & 0.024 & 0.025 \\
             & Typo         & 0.560 & 0.058 & 0.606 & 0.051 & 0.040 \\
\bottomrule
\end{tabular}
\end{table}

The shared projection has a median relative \auc{} loss of 66\% across the
18 model--family cells (range 29--95\%). Nevertheless, its difference from
random is positive on 13 cells under item-level paired 95\% bootstrap
intervals (2000 resamples, without multiplicity correction). Poetry is
numerically below random on all three checkpoints. Against own-top-16 joint
readouts, the shared projection is lower on five cells, higher on one, and
has intervals overlapping zero on twelve. This comparison is specific to
the joint readout: for residual-only recovery, shared-16 exceeds residual
own-top-16 on fourteen cells under the same interval convention.

Removing the shared directions changes joint \auc{} by $-0.063$ to $+0.141$
relative to the full readout. In particular, Mamba-2 multihop changes from
0.679 to 0.643 after removal, whereas retaining only the shared directions
gives 0.233. At rank 64, the shared projection retains more recovery, with
a median loss of approximately one third; it has no positive difference
from own-top-64 joint readouts whose interval excludes zero. Pre-update
readouts and removal of the largest pooled sensitivity mode retain the broad
geometry--recovery separation. The chosen common directions carry useful
signal but provide an incomplete semantic readout. The complement may itself
contain shared information, so this result does not identify interface-private
concepts or exclude other shared semantic constructions.

\paragraph{Interventions through the projected lenses.}
We also construct edit directions from $P_{\ell,16}J_\ell$, using the
paired two-hop and flexible-generalisation trials, $G$, $X_{\rm unit}$ and
$X_{\rm raw}$ operators, and per-site norm-matched random controls of
Section~\ref{sec:causal}. Both state-lens variants and $R/S/RS$ sites are
retained; Mamba-1 uses the same sampled suite coverage as that comparison.
Projected directions usually give success rates close to random-projector
directions. For Mamba-1 flexible generalisation, \jfull{}
$X_{\rm unit}:RS$ falls from 30/192 with the original directions to 7/192
with shared-16. A positive Mamba-3 \jfull{} $G:RS$ result, 14/192 against
2.7 successes averaged over random-projector families, is matched by
own-top-16 at 14/192. These outcomes constrain the tested edit constructions;
they do not establish that the shared subspace is unused. Simultaneous
$RS$ interventions apply both edits at full strength and do not match the
total budget of a single-interface edit.

\paragraph{Numerical qualifications.}
The local-map pilot's rule that any failed check blocks collection was not
applied to Mamba-2's sequential-scan and finite-difference tolerance misses
or Mamba-3's finite-difference misses, diagnosed as backend rounding.
Mamba-2 state-Gram eigenvalues agree with independently computed earlier
dumps to approximately $8\times10^{-4}$. Mamba-3 local maps differentiate
an exact fp32 recurrence whose forward differs from the bf16 kernel by
approximately $2\times10^{-3}$; the kernel backward had discrepancies up to
30\% at long lags. The missing BOS token in the initial Mamba-1/2 pilot
position plan was corrected before full collection. Under the dimension-scaled
fp32 rank tolerance, Mamba-2's fitted \jfull{} lens has rank below 16 at
28/63 blocks. Held-out energy retention supports the usefulness of the chosen
directions, but the fitted-lens noise floor was not estimated.

Before projection, the residual, \jfull{} and joint readouts reproduce all
saved protocol ranks under the original decoder. Projection scoring uses the equivalent
positive-rescaling-invariant RMSNorm ranking rule; fp32 near-ties change
16/7/6 of 223250/223250/82530 checked rank-capped cells for Mamba-1/2/3.
All projected and full baselines in Table~\ref{tab:shared-output-recovery}
use that same scoring path. The results describe these three checkpoints
and the specified projectors. They neither demonstrate a natural
state-to-residual transfer nor explain Mamba-2's earlier clamp response.
\FloatBarrier

\section{Implementation and numerical checks}
\label{app:implementation}

\subsection{Constructing and accumulating the state lenses}
\label{app:state-fitting}

\paragraph{Derivative paths.}
Fix a prompt, a source block and an output coordinate $j$, and suppress the
block index. Treat state vectors as columns and derivatives of the scalar
$Z_j=\sum_{u\in V_x}z_{u,j}$ as rows. Write
\begin{equation}
    r_t=\frac{\partial Z_j}{\partial y_t}C_t,
    \qquad g_t=\frac{\partial Z_j}{\partial s_t},
    \qquad A_{t+1}=\frac{\partial s_{t+1}}{\partial s_t}.
    \label{eq:state-local-derivatives}
\end{equation}
Here $A_{t+1}$ is the discrete state transition at the captured inputs,
including any coordinate rotation. The scan's input-dependent parameters are
fixed when differentiating with respect to its state: the source-block inputs
are upstream of that state. The input write and skip term therefore contribute
no additional state derivative. All paths leave $s_t$ either through its
current output $y_t$ or through $s_{t+1}$, giving the reverse recurrence
\begin{equation}
    g_{T_x-1}=r_{T_x-1},\qquad
    g_t=r_t+g_{t+1}A_{t+1}\quad(t=T_x-2,\ldots,0).
    \label{eq:state-reverse-recurrence}
\end{equation}
The $j$th rows of \jread{} and \jfull{} are respectively the fitting averages
of $r_t$ and $g_t$ in Equation~\ref{eq:lens-fit-average}. The downstream
derivative $\partial Z_j/\partial y_t$ is common to both; the control removes
only the second term of the recurrence.

\paragraph{Computing the rows.}
For each prompt, we record the final-block residual and each probed block's
scan output before gating and output projection. Backpropagating a unit
cotangent in coordinate $j$ at every valid target position simultaneously
computes $\partial Z_j/\partial y_t$. Multiplication by $C_t$ gives $r_t$,
and Equation~\ref{eq:state-reverse-recurrence} gives $g_t$. We sum each over
valid source positions, divide by $|V_x|$, and accumulate the resulting rows
in fp32 on the CPU. After all prompts, division by $N_{\rm p}$ yields the
lenses; shards are merged with weights equal to their prompt counts.
Output coordinates are processed in batches of size $b$ using replicated
prompts, requiring one forward pass and $\lceil d_{\rm model}/b\rceil$
backward passes per prompt. Residual-lens rows are captured in the same passes
as a consistency check. This procedure is implemented in
\texttt{src/fit\_state\_lens.py}, with family-specific captures in
\texttt{src/ssm\_states.py} and \texttt{src/mamba2\_states.py}.

\paragraph{Accumulation without per-token state Jacobians.}
The implementation evaluates the same source-position sum using forward
weights. For an elementwise decay $a_t$, set $w_{-1}=0$ and compute
\begin{equation}
    w_t=a_t\odot w_{t-1}+\mathbf{1}[t\in V_x],\qquad
    \sum_{t\in V_x}g_t=\sum_{t=0}^{T_x-1}w_t\odot r_t.
    \label{eq:state-weighted-sum}
\end{equation}
Each weight collects the decay products from all valid earlier source
positions. Mamba-1 uses channel- and state-coordinate-specific decays; Mamba-2
uses a scalar decay per head, broadcast over its state coordinates. The
weighted readout contractions give the required sum without storing a
Jacobian at every token. The readout-only sum instead uses the indicator
$\mathbf{1}[t\in V_x]$ directly.

\paragraph{Mamba-3 frames and MIMO readout.}
Let $\widetilde s_t=Q_t s^{\rm abs}_t$ denote the state in the current token's
rotary frame, where $Q_t$ is the corresponding orthogonal rotation. State
derivatives transform as $g^{\rm cur}_t=g^{\rm abs}_t Q_t^\top$ before
averaging. In this frame the transition combines decay with the rotation
between successive token frames; the reverse derivative applies its adjoint.
The implementation contracts readout vectors with accumulated decay and
rotary-phase weights, giving the same sum as the reverse recurrence. It also
stores an absolute-frame full-path lens as a separate comparison. For MIMO,
$y_t$ includes the readout-rank axis and $r_t$ sums the contractions over that
axis; gating and rank reduction remain in the downstream derivative. These
operations retain the same state-to-final-residual map dimensions.

\subsection{Numerical checks and verification coverage}

Exact sequential recurrences validate state capture: Mamba-1's CUDA scan
agrees with the torch loop to approximately $1.6\times10^{-6}$; Mamba-3's
kernel agrees with the fp32 reference to $2$--$4\times10^{-3}$ with bf16
inputs/outputs and fp32 state accumulation. Reverse-scan derivatives were
checked against autograd and captured residual rows against the residual-lens
estimator. At the 23 blocks shared by Mamba-2's sparse and 63-block runs,
per-block recovery \auc{} agrees to four decimal places.

The per-position release records post-update residual, state and joint ranks
at every probed block and prompt position: 541 scored prompts each for
Mamba-1/2 and 545 for Mamba-3 SISO, including 93 multihop prompts each.
Association token coverage accounts for the difference (92/92/96 of 102).
The selected temporal maps remain illustrative; population span-persistence
statistics are not reported here.

\paragraph{Verification coverage.}
Table~\ref{tab:verification} summarises automated consistency checks on the
archived results, covering summary values, result records, rank-array checksums
and arithmetic. These checks validate the covered numerical artefacts; they
do not independently replicate the model runs. Separately, all 72 recovery
\auc{} values for the conforming Pythia baseline and the Mamba-2 row-zeroed
lens were reconstructed from per-item ranks: six families, three lenses,
two layer restrictions and two runs.
\iftrue
Paired verbal-report counts at top-1 and top-5 were reconstructed from the
per-trial records, including checks of trial pairing and set unions.
\fi

\begin{table}[H]
\centering
\small
\caption{Numerical consistency checks on the archived results. All listed
checks passed. Counts refer to assertions, not independent experiments.}
\label{tab:verification}
\begin{tabular}{lr}
\toprule
Analysis checked & Assertions \\
\midrule
State and joint readouts & 974 \\
Held-out residual Jacobian geometry & 752 \\
Held-out state Jacobian geometry & 4077 \\
Readout protocols and intervention controls & 2214 \\
Direction and fixed-budget controls & 508 \\
Coordinate exchange and numerical cross-checks & 1234 \\
Paired coordinate exchange and sign-guarded steering & 3030 \\
Shared output projections and concept recovery & 2746 \\
\iftrue
Paired verbal reports: Mamba checkpoints & 782 \\
Paired verbal reports: Nemotron and Pythia & 57 \\
\fi
Workspace task suite & 4347 \\
Pursuit interventions & 1138 \\
\bottomrule
\end{tabular}
\end{table}

Three audits of saved model outputs reproduce their archived summaries exactly.
The complete two-hop audit checks 450 trial records and their evaluation
metadata; separate two-hop and flexible-generalisation audits verify
edited-concept outputs. Provenance records link tables and figures to source
artefacts, preserving SHA-256 hashes, denominators, scoring definitions and
the Mamba-3 replay qualifications. Verification is limited to the available
archived artefacts and does not establish reproducibility of every historical
experiment.

\end{document}